\documentclass[runningheads]{llncs}

\usepackage{eccv}

\usepackage{eccvabbrv}

\usepackage{multirow}
\usepackage{float}
\usepackage{adjustbox}
\usepackage{algorithm}
\usepackage{algpseudocode}
\usepackage[table]{xcolor}
\usepackage{makecell}
\usepackage{booktabs}
\usepackage{wrapfig}
\definecolor{pastelyellow}{RGB}{255, 246, 173}
\definecolor{pastelorange}{RGB}{255, 212, 130}
\definecolor{pastelred}{RGB}{255, 179, 154}

\algrenewcommand\algorithmicrequire{\textbf{Input:}}
\algrenewcommand\algorithmicensure{\textbf{Output:}}

\usepackage[pagebackref,breaklinks,colorlinks,citecolor=eccvblue]{hyperref}

\newcommand{\method}{3D FoJ}

\begin{document}

\title{3D Field of Junctions: A Noise-Robust, Training-Free Structural Prior for Volumetric Inverse Problems}

\titlerunning{3D Field of Junctions}

\author{Narges Moeini\thanks{Equal contribution; order determined by coin flip.} \and Namhoon Kim$^\star$ \and Justin Romberg \and Sara Fridovich-Keil}
\authorrunning{N. Moeini$^*$, N. Kim$^*$, J. Romberg, S. Fridovich-Keil}
\institute{Georgia Institute of Technology\\
\email{\{nmoeini3@, namhoon@, jrom@ece., sfk@\}gatech.edu}}

\maketitle

\begin{figure*}[t]
  \centering
  \includegraphics[width=\textwidth]{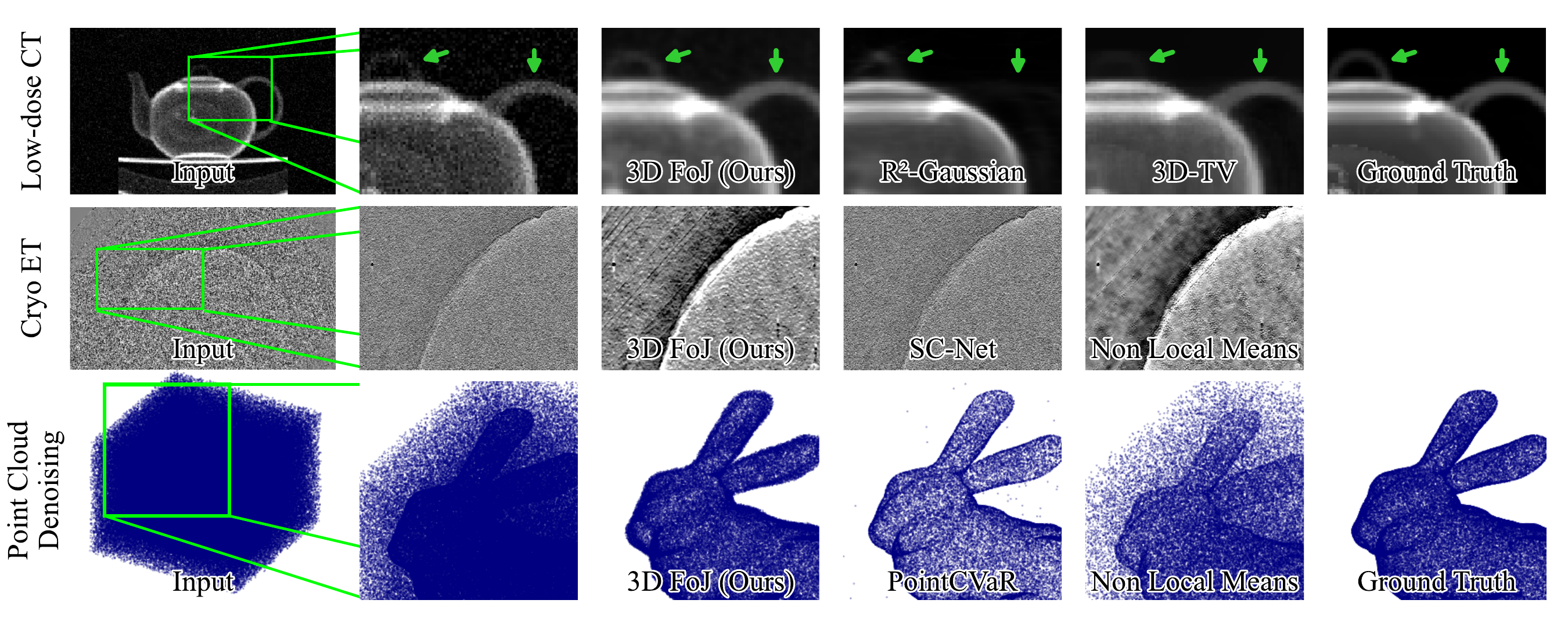}
  \caption{Our 3D Field of Junctions (\method{}) is an effective volumetric denoiser across diverse inverse problems: low-dose computed tomography (\emph{top row}), cryogenic electron tomography (\emph{middle row}), and point cloud denoising (\emph{bottom row}). Our cryo-ET experiment uses real data for which noiseless ground truth is not available. Only \method{} successfully captures both the top and side handles of the teapot in low-dose CT.}
  \label{fig:teaser}
\end{figure*}

\begin{abstract}
Volume denoising is a foundational problem in computational imaging, as many 3D imaging inverse problems face high levels of measurement noise. Inspired by the strong 2D image denoising properties of Field of Junctions (ICCV 2021), we propose a novel, fully volumetric 3D Field of Junctions (3D FoJ) representation that
optimizes
a junction of 3D wedges that best explain each 3D patch of a full volume, while encouraging consistency between overlapping patches. 
In addition to direct volume denoising, we leverage our 3D FoJ representation as a structural prior that:
(i) requires no training data, and thus precludes the risk of hallucination, (ii) preserves and enhances sharp edge and corner structures in 3D, even under low signal to noise ratio (SNR), and (iii) can be used as a drop-in denoising representation via projected or proximal gradient descent for any volumetric inverse problem with low SNR.
We demonstrate successful volume reconstruction and denoising with 3D FoJ across three diverse 3D imaging tasks with low-SNR measurements: low-dose X-ray computed tomography (CT), cryogenic electron tomography (cryo-ET), and denoising point clouds such as those from lidar in adverse weather. Across these challenging low-SNR volumetric imaging problems, 3D FoJ outperforms the evaluated classical denoisers, untrained neural denoisers, and denoisers trained only on noisy examples. Code is available at \url{https://github.com/voilalab/3D-Field-of-Junctions}.

%
\end{abstract}

\section{Introduction}

\label{sec:introduction}

Denoising is a cornerstone of signal processing. While image denoising is well-studied, volume denoising remains an active research area as many modern computational imaging modalities are plagued by low signal to noise ratio (SNR), and denoising is a key step in many iterative algorithms for volumetric inverse problems. 
In medical imaging, low SNR arises from reduced radiation dose in computed tomography (CT) \cite{lepcha2025low, smith2025projected} or weaker magnetic fields in lower-cost magnetic resonance imaging (MRI) \cite{arnold2023low}. Micro-scale electron tomography imaging uses low electron counts to prevent sample degradation \cite{turk2020promise}, while macro-scale lidar sensors experience low SNR in adverse weather \cite{yu2024modelling}.


Existing approaches to denoising often are not well-suited to 3D signals, require extensive training data, or struggle to resolve sharp boundary structure at low SNR. Classical structural and statistical priors \cite{candes2006robust, yang1980median}, and untrained neural networks like Deep Image Prior \cite{ulyanov2018deep}, Deep Decoder \cite{heckel2019deep}, and implicit neural representations \cite{tancik2020fourier, sitzmann2019siren}, can effectively denoise volumes without any training data, but tend to blur edges and corners at low SNR. Data-driven priors including diffusion models \cite{dps, dpnp} can recover sharp signal structure, but require large training datasets and are thus often restricted to learning priors over 2D images \cite{chung2023solving} or small 3D patches \cite{3dpadis} rather than full 3D volumes.
While recent volumetric reconstruction methods increasingly rely on large pretrained generative models, our work explores a complementary direction: explicit geometric modeling with strong structural inductive bias but no prior training. We demonstrate that, in extremely low-SNR regimes, such explicit structure can outperform the evaluated classical smoothness priors, untrained neural priors, and learned denoisers trained on noisy data.

Inspired by Field of Junctions (FoJ) \cite{verbin2021field}, we introduce 3D Field of Junctions (3D FoJ), a fully explicit, geometric volume representation that preserves sharp corners and surfaces while rejecting noise. Compared to the original 2D formulation \cite{verbin2021field}, 3D FoJ introduces a volumetric junction parameterization of 3D patches, an efficient implementation for large 3D volumes, and a proximal optimization framework enabling its use as a structural prior in diverse 3D inverse problems, as shown in \Cref{fig:teaser}.


\method{} works by parameterizing a volume as overlapping 3D patches, with each patch modeled by a junction consisting of (i) a set of planes that divide the patch into 3D wedges, and (ii) a constant value inside each wedge. 
By changing the positions of the planes and the values of the wedges, a 3D junction can exactly represent a sharp corner, straight or bent boundary, or constant region, and can approximate a curved boundary. Hyper-parameters control the degree to which \method{} penalizes sharp curvature.
The junction parameters are optimized through a nonconvex yet empirically stable process that begins by greedily optimizing the parameters of each 3D patch individually in parallel, and then globally refines all junction parameters jointly by encouraging plane and wedge-value consistency across overlapping 3D patches. Together, the \method{} parameterization and optimization process provide robust, boundary preserving volume denoising even under extremely low SNR.
To summarize:
\begin{itemize}
\item We introduce 3D Field of Junctions (3D FoJ), an explicit, training-free structural prior that preserves sharp geometry even under extremely low SNR.
\item We formulate 3D FoJ both as a direct volume denoiser and as a proximal regularizer for volumetric inverse problems, and show that it achieves strong performance across low-dose sparse-view CT reconstruction, cryo-ET denoising, and point cloud denoising.
\end{itemize}
Our implementation of \method{} supports single and multi-GPU parallel processing of 3D patches and chunks of 3D patches to enable reconstruction and denoising of high-resolution volumes; code will be released upon publication.

\section{Related Work}
\label{sec:relatedwork}

Many strategies have been proposed for volume denoising, broadly in the following categories.

\paragraph{Classical denoising priors.}
Classical structural priors such as bounded total variation (TV) \cite{candes2006robust}, sparsity in a transform domain \cite{yu2011dct, pierazzo2017multi}, median filtering \cite{yang1980median, zhu2012improved}, and nonlocal means filtering \cite{nonlocalmeans2005, 3Dnonlocalmeans2008, nonlocalmeansforCT2017, bm3d} can be effective denoisers for both images and volumes. These require only a few hyper-parameters and no training data, but tend to blur edges under high noise.
Our method is inspired by Field of Junctions \cite{verbin2021field}, an explicit structural prior developed for robustly extracting sharp boundary structure in 2D images at low SNR. Our work is the first to apply similar principles to low-SNR 3D inverse problems.

\paragraph{Untrained neural priors.}
Untrained neural networks such as implicit neural representations (INRs) \cite{tancik2020fourier, sitzmann2019siren, saragadam2023wire, sivgin2025geometric}, Deep Image Prior \cite{ulyanov2018deep}, and Deep Decoder \cite{heckel2019deep} leverage the architectural restrictions of a neural network as a form of implicit regularization that can remove high-frequency noise through restricting model size or training time \cite{heckel2020compressive}. 
Like their classical counterparts, untrained neural networks can be effective denoisers and require no training data, but suffer loss of boundary structure at very low SNR.

\paragraph{Data-driven neural priors.}
Trained deep network denoisers can learn complex statistical features of a training dataset, allowing them to preserve edge structure even under high noise \cite{zhu2019seismic, jia2021ddunet, ho2020denoising, zhu2023denoising}. However, such models require large training datasets of clean \cite{dps, dpnp} and/or noisy examples \cite{noise2noise, surenet}, that are often prohibitive to collect for 3D signals. While denoisers trained on 2D signals \cite{chung2023solving, chung2024decomposed, lee2023improving} or small 3D patches \cite{3dpadis} can be retroactively applied to denoise a 3D volume, these applications are approximate and some 3D structure is often lost.



\section{Preliminaries: 2D Field of Junctions}
\label{sec:preliminaries}

The Field of Junctions model~\cite{verbin2021field} provides a unified framework for representing contours, corners, and multi-way junctions in 2D images, and robustly extracts this boundary structure even under low SNR.  
FoJ models each image patch with a parametric \emph{$M$-junction}, a configuration of $M$ wedges meeting at a freely located vertex, that captures local boundary geometry within a small neighborhood.
Let $I:\Omega\!\to\!\mathbb{R}^K$ denote a $K$-channel image (\eg, $K{=}1$ for grayscale), where $\Omega \subset \mathbb{R}^2$ is the image domain.
The image $I$ is divided into overlapping patches $\{I_i\}_{i=1}^{N}$ of size $R\times R$, with stride~$s$ in each dimension.  
For the $i$th patch, the junction parameters are
\begin{equation}
\boldsymbol \theta_i = (\phi_i,\,\boldsymbol p_i^{(0)}), \qquad 
\phi_i = (\phi_i^{(1)}, \ldots, \phi_i^{(M)}),
\end{equation}
where $\phi_i^{(k)}$ denotes the orientation of the $k$th boundary, and $\boldsymbol p_i^{(0)}=(x_i^{(0)},y_i^{(0)})$ is the vertex position, which may lie inside or outside the patch.  
The number of wedges $M$ determines the local boundary complexity: for example, when $M{=}3$, the model captures three intersecting lines that can form T- or Y-junctions, but can also model straight or bent edges or uniform regions when one or more lines coincide (see \Cref{fig:foj-2d3d}(a)).    
Each wedge $j$ in junction $i$ is also endowed with a constant (optimizable) color value $c_i^{(j)} \in \mathbb{R}^K$.
The junction parameters $\Theta=\{\theta_i\}$ and colors $C=\{c_i^{(j)}\}$ are estimated by minimizing a global objective that enforces both patch fidelity and inter-patch consistency; more details are provided in \Cref{sec:appendix_2dfojoptimization} in the appendix.

\newlength{\FOJpanelH}
\setlength{\FOJpanelH}{2cm} 

\begin{figure}[t]
  \centering

  \begin{subfigure}[t]{0.62\linewidth}
    \centering
    \begin{subfigure}[b]{0.135\linewidth}
      \includegraphics[height=\FOJpanelH]{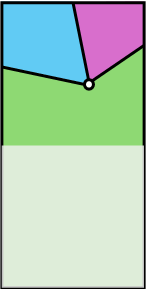}
    \end{subfigure}\hfill
    \begin{subfigure}[b]{0.135\linewidth}
      \includegraphics[height=\FOJpanelH]{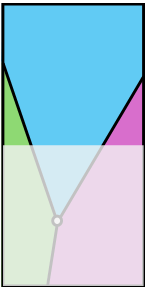}
    \end{subfigure}\hfill
    \begin{subfigure}[b]{0.135\linewidth}
      \includegraphics[height=\FOJpanelH]{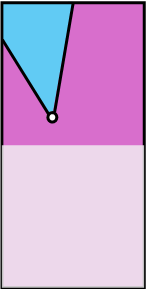}
    \end{subfigure}\hfill
    \begin{subfigure}[b]{0.135\linewidth}
      \includegraphics[height=\FOJpanelH]{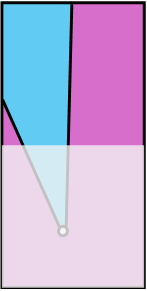}
    \end{subfigure}\hfill
    \begin{subfigure}[b]{0.135\linewidth}
      \includegraphics[height=\FOJpanelH]{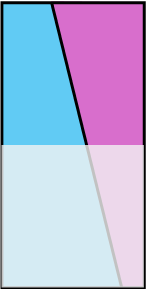}
    \end{subfigure}\hfill
    \begin{subfigure}[b]{0.135\linewidth}
      \includegraphics[height=\FOJpanelH]{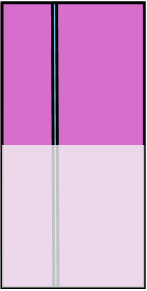}
    \end{subfigure}\hfill
    \begin{subfigure}[b]{0.135\linewidth}
      \includegraphics[height=\FOJpanelH]{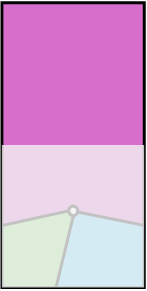}
    \end{subfigure}

    \caption{\textbf{2D FoJ} ($M{=}3$).}
    \label{fig:foj-2d}
  \end{subfigure}
  \hfill
  \begin{subfigure}[t]{0.36\linewidth}
    \centering
    \begin{subfigure}[b]{0.245\linewidth}
      \includegraphics[height=\FOJpanelH]{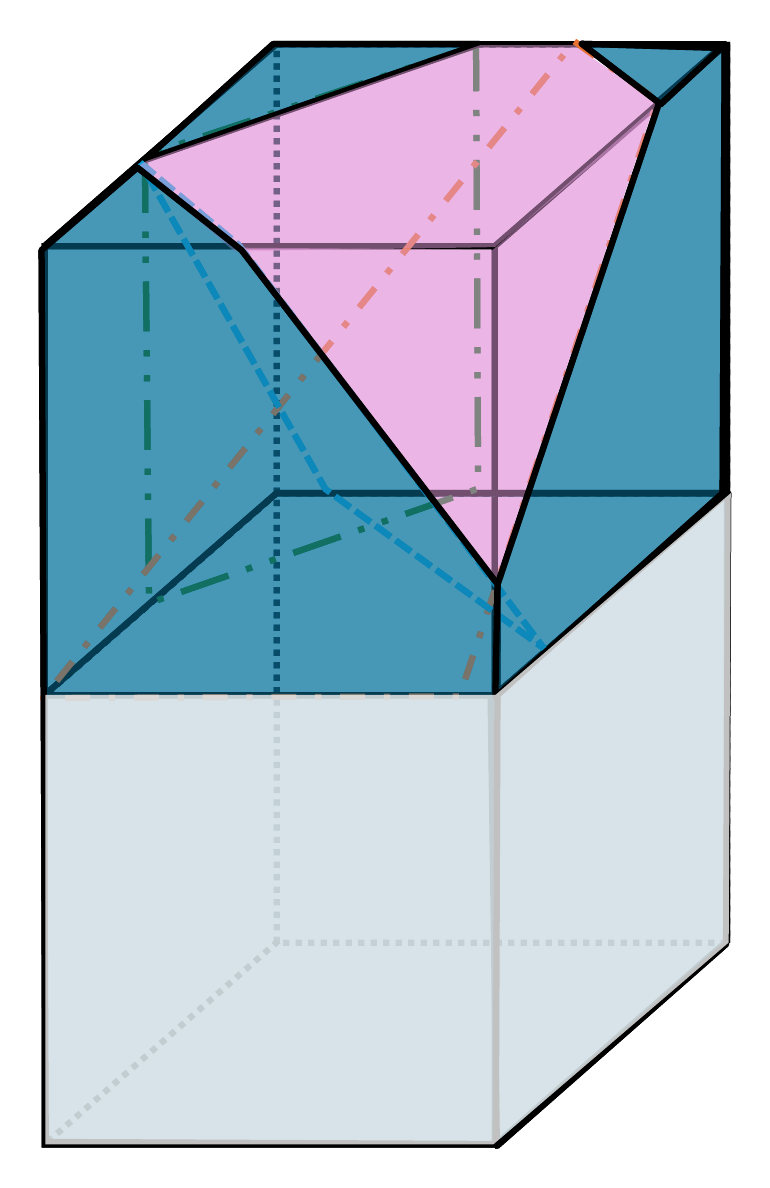}
    \end{subfigure}\hfill
    \begin{subfigure}[b]{0.245\linewidth}
      \includegraphics[height=\FOJpanelH]{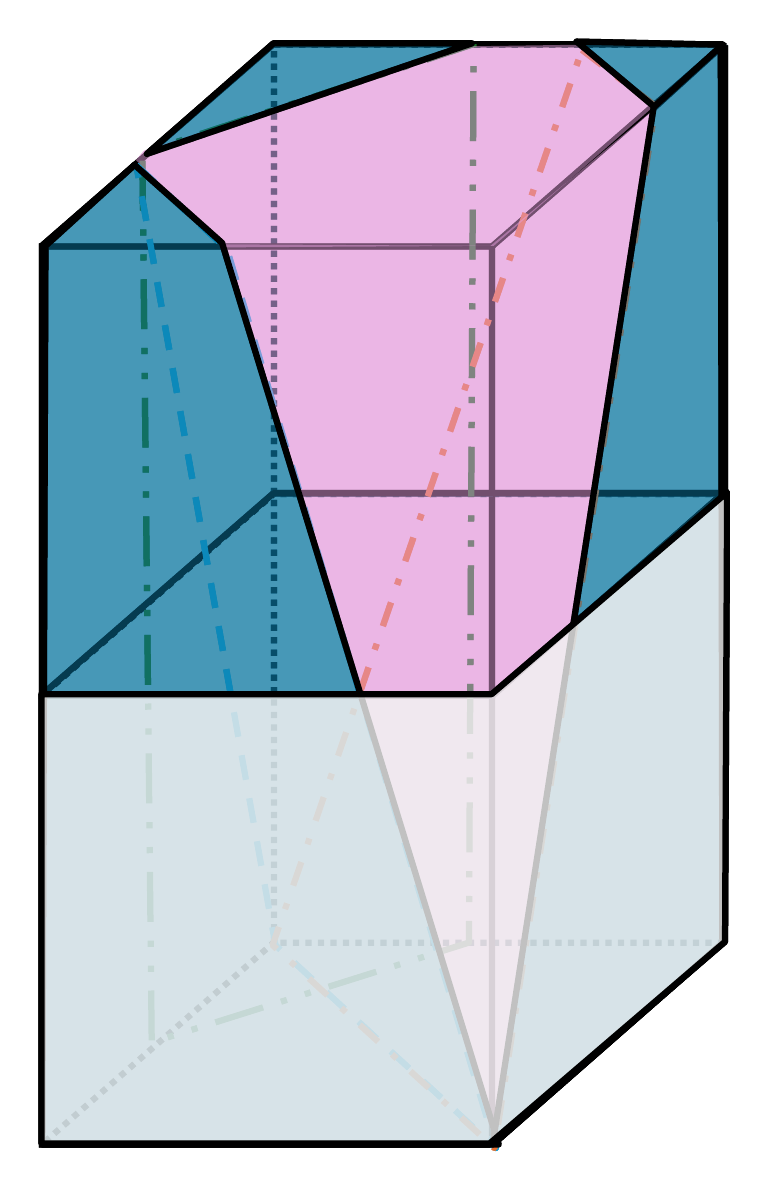}
    \end{subfigure}\hfill
    \begin{subfigure}[b]{0.245\linewidth}
      \includegraphics[height=\FOJpanelH]{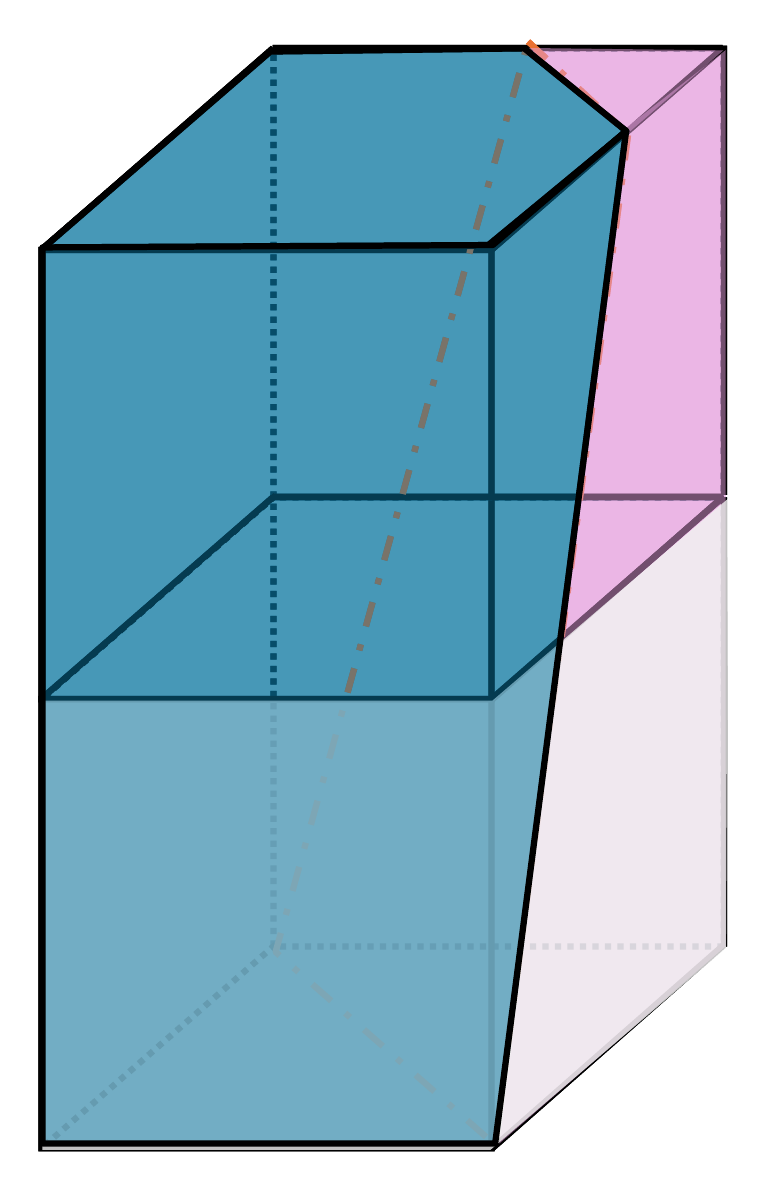}
    \end{subfigure}\hfill
    \begin{subfigure}[b]{0.245\linewidth}
      \includegraphics[height=\FOJpanelH]{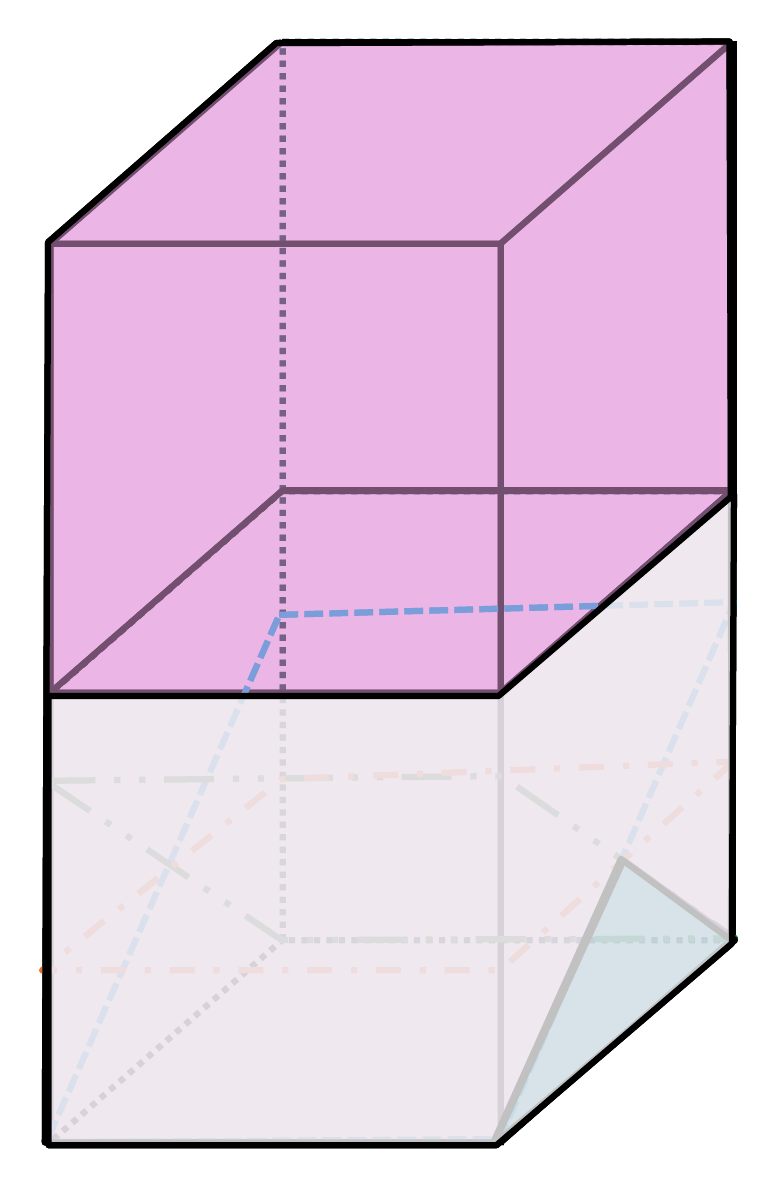}
    \end{subfigure}

    \caption{\textbf{3D FoJ} ($M{=}2$ shown).}
    \label{fig:foj-3d}
  \end{subfigure}

  \caption{\textbf{Junction schematics for Field of Junctions (FoJ) in 2D (left) and our \method{} (right).}
  \textbf{(a)} In 2D, \cite{verbin2021field} models each patch as an $M$-junction: a vertex and $M$ angles define $M$ uniform-color wedges. Allowing the vertex to lie inside or outside the patch enables a unified representation of edges, corners, junctions, and homogeneous regions.
  \textbf{(b)} In 3D, we visualize the two-region case ($M{=}2$) with different configurations of intersecting planes inside a volumetric patch. We show $M{=}2$ for visual clarity; our experiments use $M{=}3$ regions.}
  \label{fig:foj-2d3d}
\end{figure}

\section{Method: 3D Field of Junctions}
\label{sec:method}

Our \method{} is inspired by 2D Field of Junctions (FoJ)~\cite{verbin2021field} but operates natively in 3D for volumetric signals.
FoJ models image patches as junctions parameterized by intersecting lines and constant-color wedges.  
In three dimensions, we generalize these boundaries to planar surfaces that divide a 3D patch into constant-value regions. 
Like its 2D counterpart, our \method{} uses a unified junction parameterization model to represent 3D patches with many possible shapes, including
constant regions, straight and bent edges, and multi-surface junctions.  

\subsection{Local Volumetric Model}
Each volumetric patch $V_i:\Omega_i\!\to\!\mathbb{R}^K$ is modeled as a 3D region partitioned by three planes that intersect at a common vertex $\boldsymbol{v}_i^{(0)}\!\in\!\mathbb{R}^3$, which may lie inside or outside the 3D patch.  
Here, 
$\Omega_i \subset \mathbb{R}^3$ denotes the spatial domain of the 3D patch.
The full 3D volume is divided into overlapping patches $\{V_i\}_{i=1}^{N}$, each of size $R\times R\times R$, with stride~$s$ in each dimension.  
Each voxel intensity is represented as a $K$-dimensional vector (\eg, $K{=}1$ for grayscale data). The parameters of junction~$i$ are defined as
\vspace{-0.5em}
{\small \begin{equation}
\boldsymbol\gamma_i = (\boldsymbol\rho_i^{(\ell)},\,\boldsymbol{v}_i^{(0)}), ~~~~ 
\boldsymbol\rho_i^{(\ell)} = (\theta_i^{(\ell)},\phi_i^{(\ell)}), ~~\ell=1,2,3
\vspace{-0.5em}
\end{equation}}\\
where each pair $(\theta_i^{(\ell)},\phi_i^{(\ell)})$ specifies the polar and azimuthal angles of the $\ell$th plane normal, and $\boldsymbol{v}_i^{(0)}=(x_i^{(0)},y_i^{(0)},z_i^{(0)})$ denotes the vertex position at which the three planes intersect.
Since the vertex may freely move inside or outside the 3D patch, the junction can represent a wide variety of 3D structures as illustrated in \Cref{fig:foj-3d}.
The three planes divide each patch into regions (generalized 3D wedges), each of which is modeled with a constant parameter value (intensity, color, density, etc.) $c_i^{(j)} \!\in\! \mathbb{R}^K$.

\subsection{Global Objective Function}
We jointly optimize the parameters $\Gamma=(\boldsymbol\gamma_1, \dots , \boldsymbol\gamma_N)$ and regional intensities $C=(c_1, \dots , c_N)$ to minimize a global objective that enforces both local data fidelity and inter-patch coherence across the entire volume.  
Our volumetric extension of the 2D FoJ objective is given by:
{\footnotesize
\begin{equation}
\begin{aligned}
\min_{\Gamma,\,C}\quad
&\sum_{i,j}\!\!\iiint u_{\boldsymbol\gamma_i}^{(j)}(\boldsymbol{v})
\,\|V_i(\boldsymbol{v})-c_i^{(j)}\|^2\,d\boldsymbol{v}
+ \lambda_B\sum_i\!\!\iiint_{\Omega_i}
[B_i(\boldsymbol{v})-\widehat{B}(\boldsymbol{v})]^2\,d\boldsymbol{v} \\
&\quad + \lambda_C\sum_{i,j}\!\!\iiint u_{\boldsymbol\gamma_i}^{(j)}(\boldsymbol{v})
\,\|c_i^{(j)}-\widehat{V}(\boldsymbol{v})\|^2\,d\boldsymbol{v} .
\end{aligned}
\label{eq:3dfoj_objective}
\end{equation}
}

\noindent Here, $u_{\boldsymbol\gamma_i}^{(j)}(\boldsymbol{v})\!\in\!\{0,1\}$ denotes membership of a voxel $\boldsymbol{v}=(x,y,z)$ in the $j$th volumetric region of patch $i$, $V_i(\boldsymbol{v})$ represents the observed volume within patch~$i$ (with domain $\Omega_i$), and $c_i^{(j)}$ is the constant value associated with region~$j$ of the junction model for patch $i$.  
The first term enforces local reconstruction accuracy of each patch using its piecewise-constant junction model.  
The second term, weighted by $\lambda_B\!\geq\!0$, aligns local and global boundary estimates through the  boundary fields $B_i(\boldsymbol{v})$ and $\widehat{B}(\boldsymbol{v})$,
which will be defined shortly in the \textit{Global Maps} paragraph. This term encourages continuity of surface structures across overlapping patches.  
The third term, weighted by $\lambda_C\!\geq\!0$, promotes consistent regional appearance by comparing each patch’s reconstructed intensity to the global field $\widehat{V}(\boldsymbol{v})$ obtained by averaging overlapping patch reconstructions. 

\paragraph{Global Maps.}
The global volumetric color and boundary fields are defined by aggregating the contributions of all patches that contain a given voxel~$\boldsymbol{v}$.  
The global color field $\widehat{V}(\boldsymbol{v})$ is given by
{\small \begin{equation}
\widehat{V}(\boldsymbol{v})
= \frac{1}{|\mathcal{N}_{\boldsymbol{v}}|}
  \sum_{i\in\mathcal{N}_{\boldsymbol{v}}}
  \sum_{j=1}^{M}
  u_{\gamma_i}^{(j)}(\boldsymbol{v})\,c_i^{(j)},
\label{eq:3dfoj_global_color}
\end{equation}}\\
where $\mathcal{N}_{\boldsymbol{v}}$ denotes the set of all patches that include voxel~$\boldsymbol{v}$, and $|\mathcal{N}_{\boldsymbol{v}}|$ is its cardinality, the number of patches that include voxel $\boldsymbol{v}$.  
This operation computes a weighted average of all region intensities overlapping at $\boldsymbol{v}$, yielding a smooth global volume reconstruction. 
Similarly, the relaxed global boundary field $\widehat{B}(\boldsymbol{v})$ averages local boundary predictions from all overlapping patches:
{\small \begin{equation}
\widehat{B}(\boldsymbol{v})
= \frac{1}{|\mathcal{N}_{\boldsymbol{v}}|}
  \sum_{i\in\mathcal{N}_{\boldsymbol{v}}}
  B_i(\boldsymbol{v}),
\label{eq:3dfoj_global_boundary}
\end{equation}}\\
where $B_i(\boldsymbol{v})\in \{0,1\}$. This field encodes the presence or absence of a boundary plane at each voxel, effectively representing the local boundary consensus across overlapping patches. Although the binary formulation provides a clear geometric interpretation, it is not differentiable and therefore unsuitable for gradient-based optimization.   
To enable efficient and stable optimization of the junction parameters $\boldsymbol\gamma_i$, we replace the binary region indicators and boundary maps with differentiable, smooth counterparts, introduced in the following subsection.

\subsection{Smooth Indicator Functions}
\label{sec:smooth_indicator}
To enable gradient-based optimization, the region indicator functions in \Cref{eq:3dfoj_objective} must be differentiable.  
We therefore define soft indicator functions using signed distance values relative to the three separating planes of each 3D patch.  
For voxel $\boldsymbol{v}=(x,y,z)$, the signed distance to the $\ell$th plane is given by
{ \begin{equation}
d_\ell(\boldsymbol{v}) = \langle (x,y,z) - (x_i^{(0)},y_i^{(0)},z_i^{(0)}),\,\mathbf{n}_i^{(\ell)} \rangle,
\label{eq:3dfoj_distance}
\end{equation}}\\
where $\mathbf{n}_i^{(\ell)}$ is the normal vector of the $\ell$th plane, parameterized by its polar and azimuthal angles $(\theta_i^{(\ell)}, \phi_i^{(\ell)})$.
Inspired by the 2D FoJ formulation \cite{verbin2021field}, we use a Heaviside function as a differentiable approximation of the binary region assignment $ H_\eta(d) = \frac{1}{2}\!\left(1 + \frac{2}{\pi}\arctan\!\frac{d}{\eta}\right)$, where $\eta>0$ controls the transition sharpness across the plane boundaries.  
The differentiable volumetric region indicators are then defined as
{\small
\begin{align}
u_{\boldsymbol\gamma_i}^{(1)}(\boldsymbol{v}) &= H_\eta(d_1(\boldsymbol{v}))\,[1-H_\eta(d_2(\boldsymbol{v}))]\,[1-H_\eta(d_3(\boldsymbol{v}))], \nonumber\\
u_{\boldsymbol\gamma_i}^{(2)}(\boldsymbol{v}) &= [1-H_\eta(d_1(\boldsymbol{v}))]\,H_\eta(d_2(\boldsymbol{v}))\,[1-H_\eta(d_3(\boldsymbol{v}))],\nonumber\\ 
u_{\boldsymbol\gamma_i}^{(3)}(\boldsymbol{v}) &= [1-H_\eta(d_1(\boldsymbol{v}))]\,[1-H_\eta(d_2(\boldsymbol{v}))]\,H_\eta(d_3(\boldsymbol{v})),
\label{eq:3dfoj_u}
\end{align}}\\
for our parameterization with 3 regions per patch. 
By following a similar pattern and enumerating over all 8 binary arrangements with 3 planes, our 3D junction model can represent up to 8 distinct regions per 3D patch.
These functions softly assign each voxel to one of these three regions based on its position relative to the planes, providing continuous gradients for optimization.

We note that partitioning a 3D patch into $M$ disjoint regions is more geometrically complex than partitioning a 2D patch. Our \method{} formulation based on three intersecting planes naturally divides a 3D patch into eight regions; we work with fewer (typically $M=3$) regions by only assigning nonzero values to a subset of these eight regions. In particular, the region assignments in \Cref{eq:3dfoj_u} may leave some portion of the patch outside any of the $M=3$ regions and thus with default value zero. Effectively we find that these zero-valued regions do not negatively impact reconstruction or denoising quality; we use $M=3$ in our experiments to balance the expressivity of having more regions per junction with the increased computational and memory cost. Ablation studies over the choice of $M$ are provided in \Cref{tab:ablation} and \Cref{tab:ablation_appendix}. 

\paragraph{Smooth Boundary Map.}
To compare and align local surfaces across overlapping patches, we use a differentiable boundary-strength field from the 2D FoJ formulation~\cite{verbin2021field}:
{\small \begin{equation}
B_i^{(\delta)}(\boldsymbol{v})
= \pi\,\delta\, H'_{\delta}\!\Big(\min_{i\in{1,2,3}}\{\,|d_i(\boldsymbol{v})|\}\Big) \in [0,1].
\label{eq:3d_softB}
\end{equation}}

\noindent Here $H_{\delta}(d)$
is the regularized Heaviside function 
and $H'_{\delta}(d)$ is its derivative with respect to $d$. 
The input $d$ to $H'_{\delta}$ is the distance of voxel $\boldsymbol{v}$ to the nearest separating plane; thus $B_i^{(\delta)}(\boldsymbol{v})$ attains high values near predicted surfaces and smoothly decays with distance. 
The parameter $\delta>0$ controls the transition width: larger $\delta$ yields broader, softer boundaries; smaller $\delta$ produces sharper peaks. 
This soft boundary field serves as a differentiable surrogate for a binary boundary map, and is used in the boundary-consistency term of \Cref{eq:3dfoj_objective} to encourage inter-patch alignment of boundary planes.

\subsection{Optimization Procedure}

Reconstructing the volumetric Field of Junctions involves minimizing a non-convex objective over the 3D junction parameters $\boldsymbol \gamma_i = (\boldsymbol\rho_i, \boldsymbol v_i^{(0)})$.  
Similar to the 2D case \cite{verbin2021field}, we perform this nonconvex optimization in two stages: an initialization stage that estimates local geometry for each patch independently, and a refinement stage that jointly optimizes all patches using differentiable indicator functions. A detailed step-by-step version of this optimization process is provided in the Appendix. 

\vspace{-1em}
\paragraph{Initialization Step.} The initialization stage provides a coarse yet reliable estimate of the plane orientations and their vertex positions within each patch. Each junction is initialized with the vertex $\boldsymbol{v}_i^{(0)}$ placed at the center of the patch, and the plane orientation parameters $\boldsymbol\rho_i^{(\ell)} = (\theta_i^{(\ell)}, \phi_i^{(\ell)})$ are optimized using a discrete coordinate-descent search. For each orientation pair $(\theta, \phi)$, the data fidelity term from \Cref{eq:3dfoj_objective} is evaluated over a uniform angular grid, and the minimizing direction is selected. 
After the optimal plane orientations are estimated, the vertex position $\boldsymbol v_i^{(0)} = (x_i^{(0)}, y_i^{(0)}, z_i^{(0)})$ is refined by scanning a small neighborhood around its current position to minimize the same data term.  
This coordinate-wise update is repeated cyclically across the vertex and all planes until convergence, defined by a fixed number of iterations.  
This procedure is robust to noise and computationally efficient, as all patches can be processed in parallel.

\vspace{-1em}
\paragraph{Refinement Step.}
After initialization, all patches are jointly refined using gradient-based optimization of the full objective function in \Cref{eq:3dfoj_objective}. The binary region indicators and boundary maps are replaced by their differentiable counterparts (defined in Section~\ref{sec:smooth_indicator}), enabling end-to-end optimization through the junction parameters $\boldsymbol\gamma_i$. We use first-order iterative optimization (e.g., Adam) to update $\theta_i^{(\ell)}$, $\phi_i^{(\ell)}$, and $\boldsymbol v_i^{(0)}$, while the region intensities $c_i^{(j)}$ are updated analytically at each iteration using the following closed-form solution:
{\small \begin{equation}
c_i^{(j)} =
\frac{\displaystyle \iiint u_{\boldsymbol\gamma_i}^{(j)}(\boldsymbol{v})\,[V_i(\boldsymbol{v})+\lambda_C\,\widehat{V}(\boldsymbol{v})]\,d\boldsymbol{v}}
{\displaystyle (1+\lambda_C)\!\iiint u_{\boldsymbol\gamma_i}^{(j)}(\boldsymbol{v})\,d\boldsymbol{v}}.
\label{eq:3dfoj_sigma_hard}
\end{equation}}\\
During refinement, the regularization weights $\lambda_B$ and $\lambda_C$ are gradually increased from small initial values to their final targets, ensuring stable convergence and avoiding premature over-smoothing.  
This joint optimization progressively improves both the geometric accuracy of the surfaces and their spatial consistency across overlapping patches, yielding a coherent 3D Field of Junctions.

\subsection{Solving Inverse Problems with 3D FoJ}
Our \method{} model can be used as a drop-in structural prior in any volumetric inverse problem; we showcase it in the context of low-dose sparse-view tomographic reconstruction.
The reconstruction problem is formulated as

\begin{equation}
\min_{\boldsymbol x}\; f(\boldsymbol x)+g(\boldsymbol x), 
\label{eq:foj_inverse_objective}
\end{equation}\\
where $\boldsymbol x\!\in\!\mathbb{R}^{D\times H\times W\times K}$ denotes the volumetric image to be reconstructed. The data fidelity term $f(\boldsymbol{x})$ encourages consistency with the measured projections:
\begin{equation}
f(\boldsymbol{x})=\tfrac{1}{2}\|\boldsymbol{A}\boldsymbol{x}-\boldsymbol{b}\|^2,
\end{equation}
where $\boldsymbol{A}$ is the system matrix (for tomography, the Radon transform) and $\boldsymbol{b}$ is the vector of projection measurements.  
The \method{} regularization term
\begin{equation}
g(\boldsymbol{x})=\min_{\Gamma, C}\|\boldsymbol{x}-R(\Gamma, C)\|^2
\end{equation}
encourages geometric consistency in voxel space, with $R(\Gamma, C)$ denoting the reconstructed volume parameterized by \method{} parameters~$\Gamma$ and \textit{C}.  
We solve \Cref{eq:foj_inverse_objective} via a proximal-gradient method:

\begin{equation}
\boldsymbol x^{(k+1)}=\mathrm{prox}_{\lambda g}\!\left(\boldsymbol x^{(k)}-\lambda\nabla f(\boldsymbol x^{(k)})\right),
\label{eq:foj_pgd}
\end{equation}
which alternates between a gradient descent step on the data term and a proximal step defined by the FoJ prior. The scalar $\lambda>0$ serves a dual role: it acts as the gradient descent step size on the data fidelity term 
$f(\boldsymbol x)$, and simultaneously controls the strength of the \method{} regularization through the proximal operator. 
As $\lambda \!\downarrow\! 0$, the update approaches a pure gradient descent step with negligible \method{} influence, and the reconstruction converges toward the conventional least-squares solution. Conversely, as $\lambda \!\uparrow\! \infty$, the proximal step dominates and the solution is driven toward the \method{} manifold, yielding reconstructions that closely follow the \method{} structural prior $R({\Gamma, C})$. 
In practice, $\lambda$ is tuned to balance data fidelity and \method{} regularization. The two updates can be written explicitly as:
\begin{align}
\boldsymbol x^{(k+\frac{1}{2})}
&= \boldsymbol x^{(k)} - \lambda\,\nabla f\!\big(\boldsymbol x^{(k)}\big).
\label{eq:cgls_step} \\
\boldsymbol x^{(k+1)} &=
\arg\min_{\boldsymbol{x}}\Bigl[g(\boldsymbol x)+\tfrac{1}{2\lambda}\,\|\boldsymbol{x}-\boldsymbol{x}^{(k+\frac{1}{2})}\|^{2}\Bigr].
\label{eq:foj_step1}
\end{align}
The first half-iteration (\Cref{eq:cgls_step}) corresponds to a conventional gradient descent update enforcing data consistency,
and the second half-iteration (\Cref{eq:foj_step1}) corresponds to a proximal update to encourage consistency with the denoised 3D FoJ representation.
In practice, we approximate this proximal step efficiently by performing a single 3D FoJ update (one iteration of the initialization step and one iteration of the refinement step) per outer iteration, initialized from the intermediate volume. 
Empirically, we find that a single FoJ update at each proximal step is sufficient to guide the solution toward the FoJ manifold while keeping the overall reconstruction computationally tractable. 
This proximal formulation enables joint recovery of the voxel intensity field and the underlying junction geometry within each iteration, using noisy tomographic measurements rather than direct access to a noisy volume.
We emphasize that, while we present results on a tomographic inverse problem, the same proximal gradient formulation would enable \method{} to be used as a drop-in structural prior to regularize any volumetric inverse problem.
We also note that, while the \method{} optimization process is nonconvex, empirically we observed stable optimization across all experiments. Empirical update-norm convergence curves for the proximal reconstruction procedure are provided in \Cref{sec:app_convergence}.

\section{Results}
\label{sec:results}

To illustrate the broad applicability of \method{} as a structural prior for volume reconstruction and denoising, we perform experiments on three diverse volume imaging tasks that share the key feature of inherent measurement noise: low-dose X-ray computed tomography (CT), cryogenic electron tomography (cryo-ET), and point cloud denoising. Low-dose CT and cryo-ET face severe shot noise due to low X-ray and electron beam intensities, which are restricted to avoid radiative damage and sample degradation, respectively. Point clouds reconstructed from sensors such as lidar can be corrupted by outliers due to sensor noise or adverse weather. Our experiments frame X-ray CT as an inverse problem, where we use \method{} inside proximal gradient descent to reconstruct a denoised volume from raw noisy projections. For cryo-ET and point cloud denoising, we start with a noisy volume and showcase \method{} as a direct 3D denoiser.

  

\subsection{Low-dose X-ray Computed Tomography (CT)}

\begin{wrapfigure}{r}{0.6\columnwidth}
 \vspace{-2.3em}
\centering
\setlength{\tabcolsep}{0pt}
\begin{tabular}{@{}c@{}}
  \includegraphics[width=\linewidth]{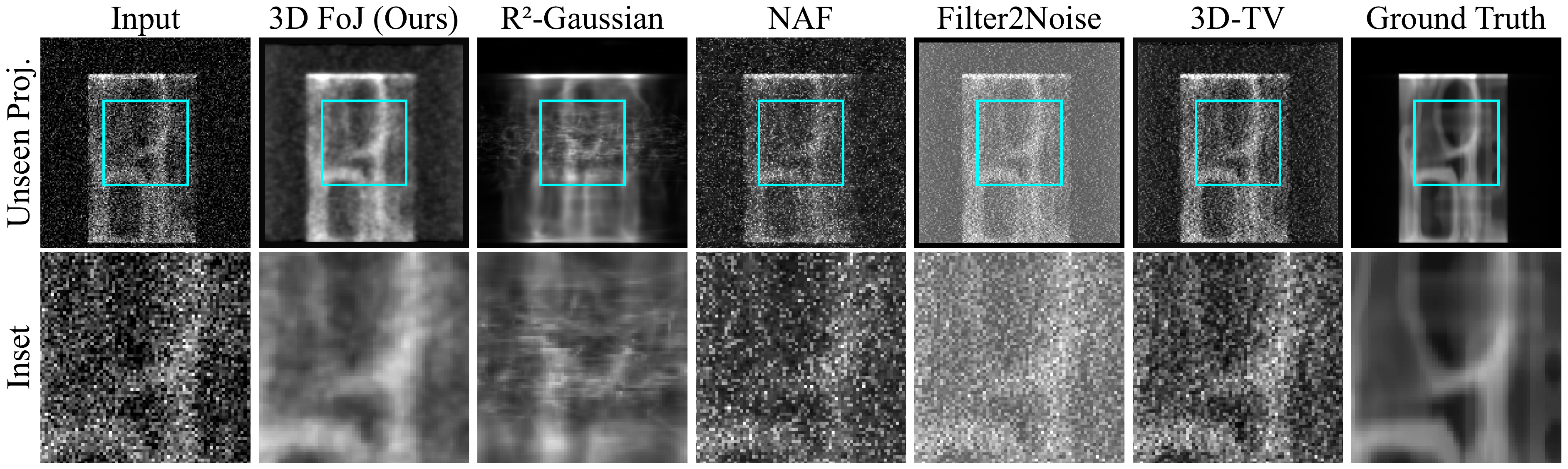}\\[-0.4em]
  \includegraphics[width=\linewidth]{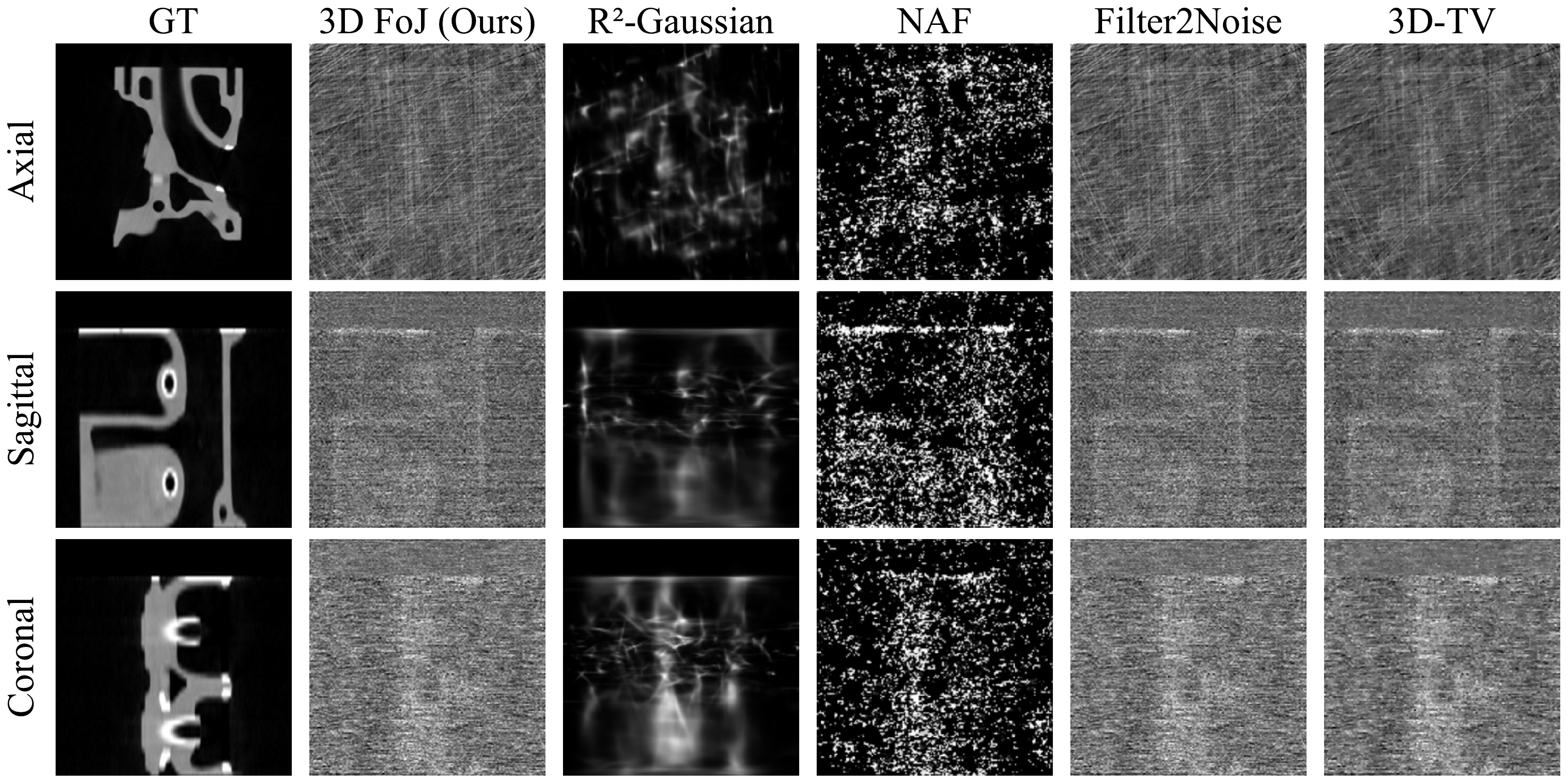}
\end{tabular}
\caption{\footnotesize Comparison of unseen projection views synthesized from reconstructed 3D volumes (top) and slice views of reconstructed 3D volumes (bottom) for the \textit{engine} dataset under extremely low-SNR conditions (\textit{P50}, severe shot noise with only 50 photons per detector pixel).}
\label{fig:ct_engine_p50_proj_slice}
\vspace{-2.5em}
\end{wrapfigure}

To evaluate the reconstruction performance of our \method{} for limited-angle and low-dose 3D CT reconstruction, we conduct experiments on five synthetic volumetric datasets—\textit{pepper, teapot, jaw, foot,} and \textit{engine}—originally introduced in the R$^2$-Gaussian paper~\cite{r2_gaussian}.  
We reconstruct each object at resolution $256^3$ from 20 non-uniform projection views, each with resolution $256^2$.  
To emulate low-dose X-ray acquisition, we add Poisson-distributed noise to the synthetic projection data, where the simulated photon count per detector pixel determines the effective signal-to-noise ratio (SNR).

We simulate three photon-count levels, denoted as \textit{P50}, \textit{P100}, and \textit{P1000}, corresponding respectively to increasing SNR regimes.  
Here, the notation \textit{P}$n$ indicates an expected photon count of $n$ per pixel (in air), with \textit{P50} representing the lowest SNR (extreme noise) and \textit{P1000} representing the highest SNR (but still noisy) condition. Additional qualitative comparisons of unseen projection views (\Cref{fig:ct_p50_results_appendix,fig:ct_p100_results_appendix,fig:ct_p1000_results_appendix}) and reconstructed axial, sagittal, and coronal slice views (\Cref{fig:ct_p50_slice_results_appendix,fig:ct_p100_slice_results_appendix,fig:ct_p1000_slice_results_appendix}) across all datasets and noise levels further demonstrate the consistent structural fidelity and noise robustness of \method{}.


\newcommand{\rotmetric}[1]{%
  \makebox[0pt][l]{\hspace{1.2em}\rotatebox{90}{\scriptsize\bfseries #1}}%
}

\newcommand{\rotmetricbox}[1]{%
  \rotatebox{90}{\parbox[c][1.55cm][c]{1.55cm}{\centering\scriptsize\bfseries #1}}%
}

\begin{wraptable}[14]{h}{0.450\columnwidth}
\vspace{-2.3em}
\centering
\scriptsize
\setlength{\tabcolsep}{2.4pt}
\renewcommand{\arraystretch}{1.0} 

\resizebox{\linewidth}{!}{%
\vspace{1cm}
\begin{tabular}{c lccc}
\toprule
\textbf{Metric} & \textbf{Method} & \textbf{P50} & \textbf{P100} & \textbf{P1000} \\
\midrule
\multirow{5}{*}{\rotatebox{90}{\scriptsize\textbf{\shortstack{2D\\MS\text{-}SSIM}}}}
& 3D-TV & 0.703 & 0.739 & 0.829 \\

& Filter2Noise & 0.663 & 0.689 & 0.695 \\
& NAF & 0.423 & 0.467 & 0.569 \\
& $\mathrm{R}^2$-Gaussian & 0.574 & 0.577 & 0.652 \\
& \method{} (ours) & \cellcolor{pastelyellow}{0.754} & \cellcolor{pastelyellow}{0.788} & \cellcolor{pastelyellow}{0.838} \\
\midrule
\multirow{5}{*}{\rotatebox{90}{\scriptsize\textbf{\shortstack{3D\\PSNR}}}}
& 3D-TV & 13.79 & 15.31 & 16.96 \\

& Filter2Noise & 11.61 & 15.13 & 14.64 \\
& NAF & 14.17 & 14.67 & 15.66 \\
& $\mathrm{R}^2$-Gaussian & 15.29 & 15.29 & 15.87 \\
& \method{} (ours) & \cellcolor{pastelyellow}{15.88} & \cellcolor{pastelyellow}{16.14} & 16.72 \\
\bottomrule
\end{tabular}%
}
\vspace{-3.3mm}
\caption{\scriptsize Average MS-SSIM on 2D projections (360 views) and 3D PSNR on volumes across all CT datasets (\emph{engine, foot, jaw, pepper}, and \emph{teapot}) for each noise level. The best method is highlighted for each noise level and metric.
}
\vspace{-5em}
\label{tab:combined_PSNR_MSSSIM}
\end{wraptable}

We compare our \method{} reconstructions against baselines that do not require paired clean/noisy 3D training data: an instance-optimized explicit representation, R$^2$-Gaussian~\cite{r2_gaussian}; an instance-optimized neural representation, Neural Attenuation Fields (NAF)~\cite{zha2022naf}; a self-supervised denoiser trained only on noisy examples, 
Filter2Noise~\cite{filter2noise}; and a classical volumetric regularizer, 3D total variation (TV) regularization~\cite{TV}. Fully supervised volumetric denoisers trained with matched clean/noisy CT data are outside our evaluation setting, as are large generative priors that require external 2D or patch-based training. 
Our goal is to compare against methods that can be applied at the full-volume scale under similar computational budgets and without an external training set of paired clean and noisy volumes.

We evaluate reconstruction fidelity using both volumetric and projection-based metrics. Specifically, we report the 3D peak signal-to-noise ratio (3D PSNR) computed on the reconstructed volumes and the multi-scale structural similarity (MS-SSIM) \cite{MS-SSIM} measured on 360 2D projections (one per degree along a circular trajectory).
\Cref{tab:combined_PSNR_MSSSIM} summarizes the average MS-SSIM and 3D PSNR across all datasets (\textit{engine}, \textit{foot}, \textit{jaw}, \textit{pepper}, and \textit{teapot}) and noise levels.
As shown, \method{} achieves the strongest average structural fidelity across all noise levels, with particularly clear gains under low-SNR (low photon count) conditions. 
For volumetric PSNR, \method{} performs best at P50 and P100 and remains competitive at P1000. 
\Cref{fig:ct_engine_p50_proj_slice} provides qualitative comparisons of the reconstructed geometry under the low-SNR (\textit{P50}) condition. The top row shows an unseen projection synthesized from each reconstructed 3D volume, which tests global consistency for a view not included in the input angles. The bottom row shows slice views of the reconstructed volumes, where \method{} preserves cleaner surfaces, sharper edges, and more faithful structural details than all baselines. 
The full quantitative comparison across all datasets and noise levels, including both projection-based MS-SSIM and volumetric 3D PSNR metrics, is provided in \Cref{tab:combined_PSNR_MSSSIM_appendix} in the appendix.


\vspace{-1em}
\subsection{Cryogenic Electron Tomography (cryo-ET)}

To evaluate the denoising capability of our method on real-world electron microscopy data, we conducted experiments on four cryo-ET datasets: \emph{centriole} from \cite{kremer1996imod} and \emph{mitochondria}, \emph{vesicle}, and \emph{VEEV} from \cite{cryoet2021}. 
We take as input the noisy $1024\times1024\times300$ volumes reconstructed using 
the \texttt{tilt} program in IMOD~\cite{kremer1996imod} as described in the SC-Net pipeline~\cite{cryoet2021}. 
Noise in the input volumes is the result of both electron shot noise, because the electron beam intensity is limited to avoid sample degradation, and missing wedge artifacts, because cryo-ET projections are geometrically constrained to measure limited angles.




\begin{figure}
\vspace{-1.5em}
\centering
\includegraphics[width=0.85\linewidth]{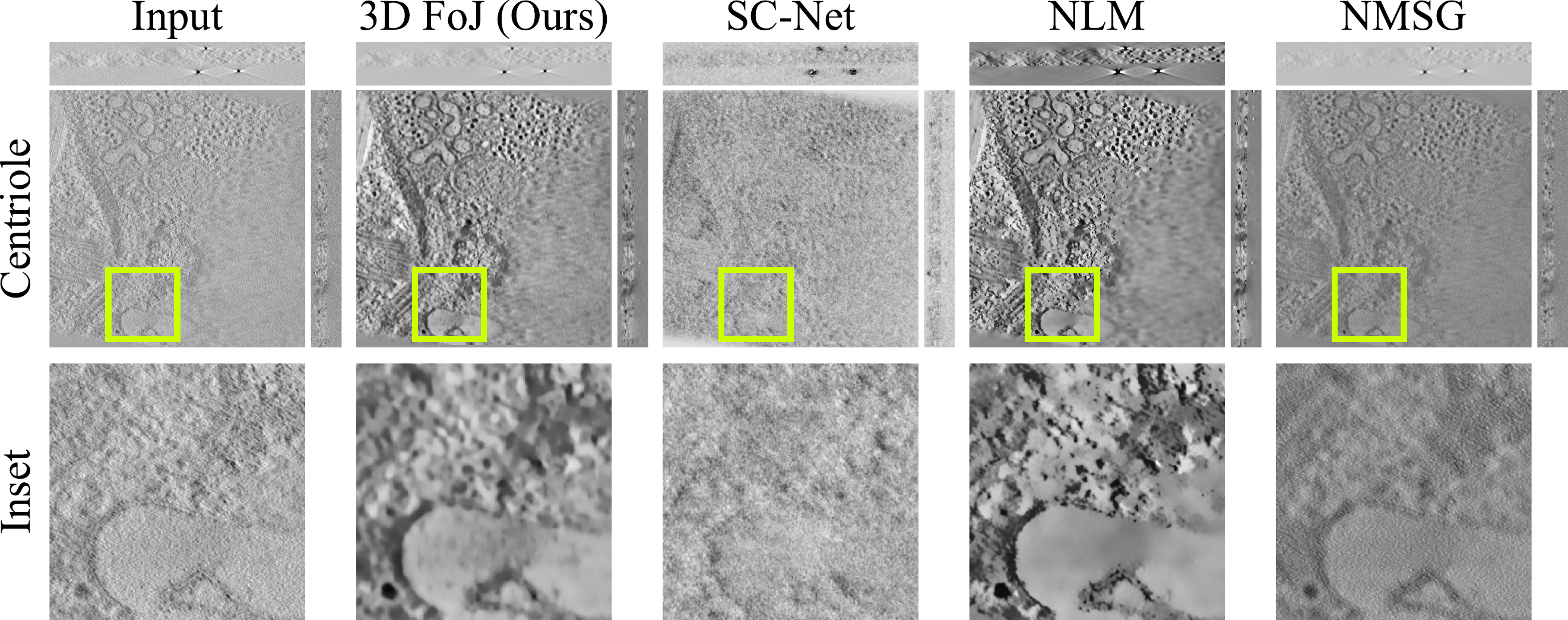}
\vspace{-0.5em}
\caption{\footnotesize Visual denoising results for the cryo-ET \emph{centriole} volume, sliced in each dimension. \method{} preserves structural details that are overly smoothed by other methods.}

\label{fig:cryoet_results}
\vspace{-2.em}
\end{figure}

Since noiseless ground-truth volumes are unavailable for real cryo-ET data, we use Fourier Shell Correlation (FSC)~\cite{harauz1986FSC} for quantitative evaluation, following standard practice in cryo-ET reconstruction. 
FSC measures the frequency-domain consistency between independently reconstructed subsets of the available projections. 
For the IMOD \emph{centriole} dataset, we split projections into even and odd halves following SC-Net~\cite{cryoet2021}, reconstruct both halves in IMOD, denoise each half independently, and compute the half-map FSC $\rho_{\mathrm{half}}(k)$ between the two denoised half-maps. 
We report the corrected even/odd FSC curves using the correction adopted from \cite{cryoet2021}:
$\mathrm{FSC}_{\mathrm{e/o}}(k)=2\rho_{\mathrm{half}}(k)/(1+\rho_{\mathrm{half}}(k))$.
Under this correction, the equivalent of the gold-standard $0.143$ half-map FSC threshold is $0.25$.

We compared our \method{} method against SC-Net~\cite{cryoet2021}, Noise Modeling and Sparsity Guidance (NMSG)~\cite{yang2024nmsg}, and nonlocal means denoising (NLM)~\cite{nonlocalmeansforCT2017} as representative neural~\cite{cryoet2021,yang2024nmsg} and classical~\cite{nonlocalmeansforCT2017} denoisers applicable to cryo-ET volumes.

\begin{figure}[h]
\vspace{-2.em}
  \centering
  \begin{minipage}[c]{0.50\linewidth}
    \centering
    \includegraphics[width=0.9\linewidth]{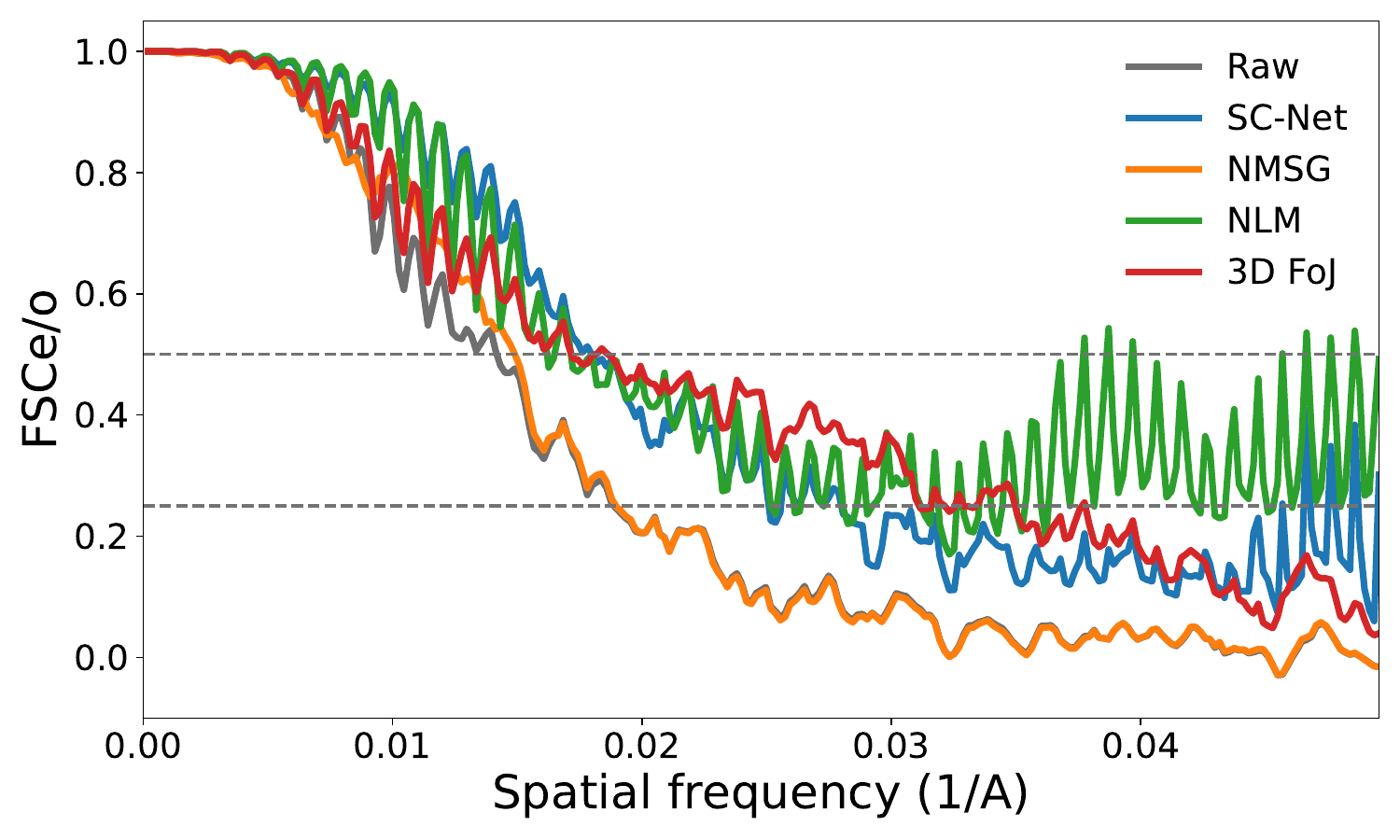}
  \end{minipage}
  \hfill
  \begin{minipage}[c]{0.49\linewidth}
    \centering
    \setlength{\tabcolsep}{2.5pt}
    \renewcommand{\arraystretch}{0.92}
    \scriptsize
    \begin{tabular}{lcc}
    \toprule
    & \multicolumn{2}{c}{Resolution (\AA)\,$\downarrow$} \\
    Method & \shortstack{FSC$_{\text{e/o}}$\\$0.5$} & \shortstack{FSC$_{\text{e/o}}$\\$0.25$} \\
    \midrule
    SC-Net & \textbf{55.60} & 39.86 \\
    NMSG   & 67.10 & 52.76 \\
    NLM    & 61.73 & 39.75 \\
    3D FoJ & 58.24 & \textbf{32.14} \\
    \bottomrule
    \end{tabular}
  \end{minipage}
  \vspace{-2mm}
  \caption{\footnotesize Even/odd FSC evaluation on the IMOD \emph{centriole} dataset. 
  }
  \label{fig:FSC}
  \vspace{-5mm}
\end{figure}

As shown in \Cref{fig:FSC}, \method{} achieves the best FSC$_{\text{e/o}}$ resolution at the 0.25 threshold and remains competitive with SC-Net at the 0.5 threshold, while requiring no training data.
Qualitative volume denoising results in \Cref{fig:cryoet_results} show that \method{} achieves the best visual balance of denoising, contrast enhancement, and detail preservation.
Further qualitative results on additional real cryo-ET volumes are presented in the appendix (\Cref{fig:cryoet_results_appendix}), illustrating the competitive denoising performance of \method{} relative to alternative approaches.

\subsection{Point Cloud Denoising}

\begin{figure}[h]
\vspace{-2.7em}
  \centering
    \includegraphics[width=\linewidth]{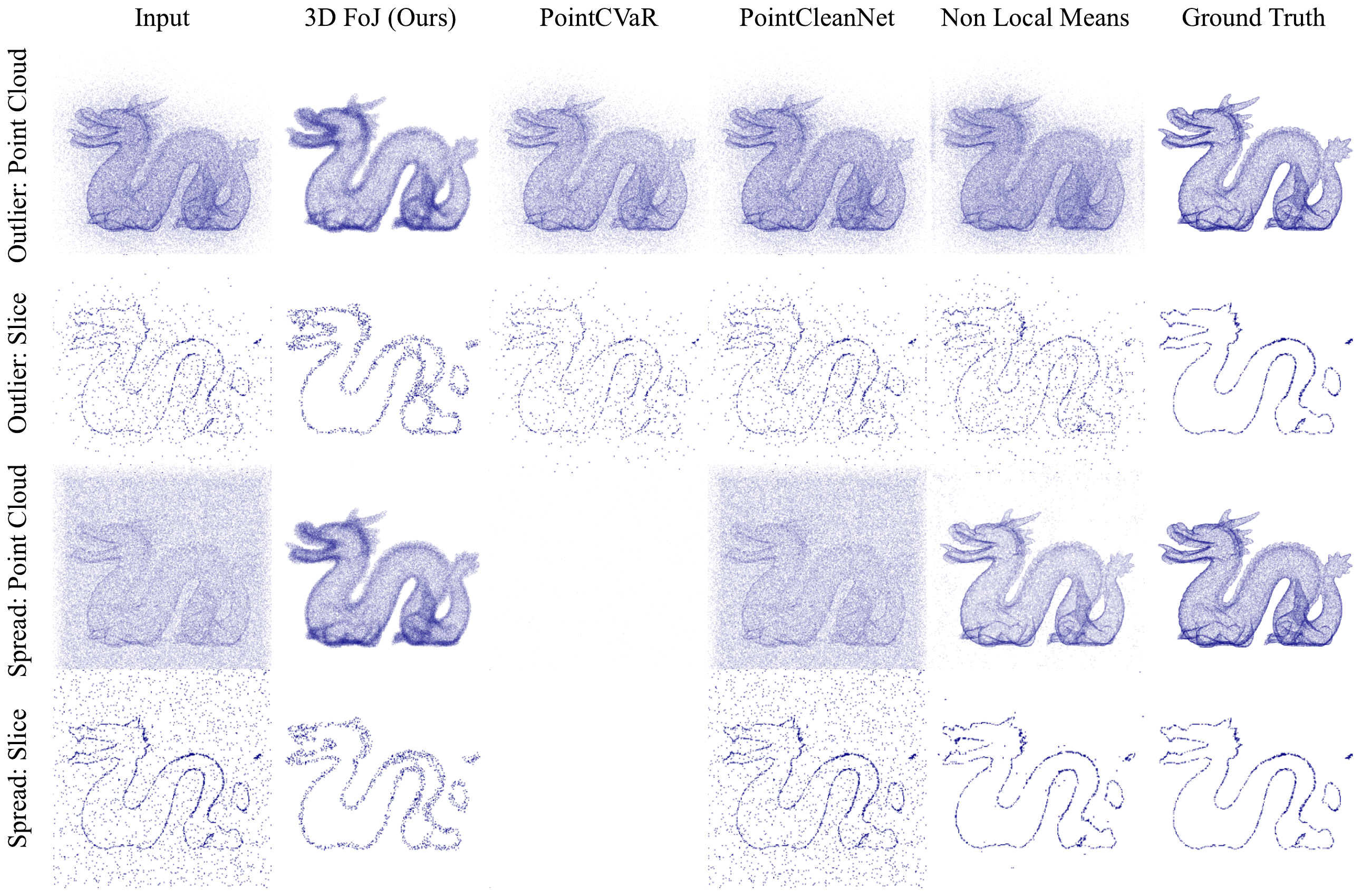}
    \vspace{-2em}
  \caption{
    Denoising results for the \emph{dragon} point cloud with 60\% \emph{outlier} noise and 500k random \emph{spread} points.
    Slices show that \method{} effectively removes noise both inside and outside the object surface. At this extreme level of \emph{spread} noise, PointCVaR rejects nearly all points as outliers.
  }
  \label{fig:spread_noise_visualization}
  \vspace{-2.em}
\end{figure}

We further evaluate our method on point cloud denoising tasks, using a total of 28 point clouds provided by PointCleanNet~\cite{rakotosaona2020pointcleannet}. 
We consider two noise models: (i) outlier noise modeled on what might be expected from lidar sensor noise or miscalibration, in which noise points are concentrated near object surfaces, and (ii) synthetic real-world noise similar to what would be expected from a lidar sensor in rain or snow. We refer to the first noise model as \emph{outlier} and the second as \emph{spread}.
Our \emph{outlier} removal experiments use the same noise configurations as PointCleanNet \cite{rakotosaona2020pointcleannet}, focusing on the four outlier ratios (10\%, 30\%, 60\%, 90\%) in their dataset. 
Our \emph{spread} denoising experiments use simulated inclement-weather lidar noise modeled on a similar experiment from PointCleanNet.
For each point cloud, we randomly add 40k, 100k, 200k, or 500k noisy points within an axis-aligned bounding box, defined by padding along each of the $x$, $y$, and $z$ axes by 10 meters.

We compare our \method{} method against PointCleanNet~\cite{rakotosaona2020pointcleannet}, PointCVaR~\cite{li2024pointcvar}, and nonlocal means (NLM)~\cite{nonlocalmeansforCT2017} as representative strong and diverse baselines for point cloud denoising. 
PointCleanNet uses a neural network for denoising, PointCVaR follows a statistical approach to reject outlier points, and nonlocal means averages structurally similar but disjoint 3D patches to reduce noise. 
Because \method{} and NLM are configured for voxel rather than point cloud denoising, for these methods we convert the noisy input point cloud into a $256^3$ voxel grid before denoising.
After denoising, we convert the denoised voxel grids from these methods back into point clouds by placing points at the highest-intensity 100k voxel coordinates, to match the point-based denoisers (PointCleanNet and PointCVaR) which are configured to output 100k points. For fairness, all voxel-based methods use the same voxelization and de-voxelization protocol, with full details provided in \Cref{point_cloud_denoising_appendix}. 

\begin{wraptable}[9]{h}{0.62\columnwidth}
\vspace{-2.2em}
\centering
\scriptsize
\setlength{\tabcolsep}{2.2pt}
\renewcommand{\arraystretch}{1.05}
\resizebox{\linewidth}{!}{%
\begin{tabular}{c lcccc}
\toprule
\textbf{Task} & \textbf{Level} & \textbf{PointCVaR} & \textbf{PointCleanNet} & \textbf{NLM} & \textbf{\method{} (Ours)} \\
\midrule
\multirow{4}{*}{\rotatebox{90}{\textbf{Outlier}}}
& 10\% & 2.15$\,\pm\,$6.47 & 2.17$\,\pm\,$6.44 & 5.02$\,\pm\,$16.73 & \cellcolor{pastelyellow}0.15$\,\pm\,$0.40 \\
& 30\% & 6.54$\,\pm\,$19.39 & 6.52$\,\pm\,$19.25 & 7.82$\,\pm\,$24.15 & \cellcolor{pastelyellow}0.16$\,\pm\,$0.42 \\
& 60\% & 12.93$\,\pm\,$38.26 & 12.98$\,\pm\,$38.19 & 13.41$\,\pm\,$40.62 & \cellcolor{pastelyellow}0.22$\,\pm\,$0.59 \\
& 90\% & 19.64$\,\pm\,$57.82 & 19.57$\,\pm\,$57.56 & 18.35$\,\pm\,$54.88 & \cellcolor{pastelyellow}1.74$\,\pm\,$5.68 \\
\midrule
\multirow{4}{*}{\rotatebox{90}{\textbf{Spread}}}
& 40k  & 154.41$\,\pm\,$396.20 & 62.33$\,\pm\,$191.76 & 87.05$\,\pm\,$279.53 & \cellcolor{pastelyellow}0.16$\,\pm\,$0.41 \\
& 100k & \cellcolor{pastelorange}0.44$\,\pm\,$0.66 & 108.50$\,\pm\,$333.13 & 138.14$\,\pm\,$438.95 & \cellcolor{pastelyellow}0.17$\,\pm\,$0.44 \\
& 200k & 256.72$\,\pm\,$661.88 & 144.94$\,\pm\,$445.71 & 168.08$\,\pm\,$527.55 & \cellcolor{pastelyellow}0.18$\,\pm\,$0.48 \\
& 500k & 515.25$\,\pm\,$1327.14 & 181.12$\,\pm\,$557.22 & 102.84$\,\pm\,$334.54 & \cellcolor{pastelyellow}0.48$\,\pm\,$1.79 \\
\bottomrule
\end{tabular}%
}
\vspace{-1em}
\caption{\scriptsize Point cloud denoising (Chamfer-L2), mean$\pm$std across shapes. Best per noise level is highlighted. PointCVaR is competitive at moderate spread noise but is less robust across noise levels.}
\label{tab:combined_ChamferL2}
\vspace{-2.5em}
\end{wraptable}
We report Chamfer Distance (CD)~\cite{levandowsky1971chamferdistance,achlioptas2018learningchamferdistance} in \Cref{tab:combined_ChamferL2} to quantify geometric consistency between denoised and ground truth point clouds.
We find that \method{} is a robust point cloud denoiser for both types of noise and across all noise levels, extracting robust shape structure even under low SNR. 
Visual results are shown in \Cref{fig:spread_noise_visualization}.
Additional quantitative evaluations under outlier noise and spread noise settings are reported in the appendix (\Cref{tab:outlier_ChamferL2_2x2_levels,tab:spread_ChamferL2_2x2_levels}), further demonstrating the robustness and consistency of \method{} across a wide range of corruption levels.

\subsection{Ablation Study}
\begin{wraptable}{r}{0.6\columnwidth}
\vspace{-5.8em}
\centering
\scriptsize
\setlength{\tabcolsep}{2.2pt}
\renewcommand{\arraystretch}{1.05}

\resizebox{\linewidth}{!}{%
\begin{tabular}{c c c c c c c}
\toprule
\textbf{\# regions}& \textbf{Batch size} & \textbf{Stride} & \textbf{Patch size} &  \textbf{MS-SSIM} & \textbf{3D PSNR} & \textbf{Runtime}\\
\midrule
3&6&1&4 & 0.755 & 11.84 & 14.3 min\\
3&6&1&6 & 0.732 & 14.29 & 31.6 min\\
3&6&1&8 & 0.722 & 15.76 & 47.3 min\\
3&6&2&4 & 0.762 & 11.33 & 7.2 min\\
3&6&2&6 & 0.784 & 14.37 & 10.5 min\\
3&6&2&8 & 0.796 & 16.18 & 13.1 min\\
3&6&2&10& 0.798 & 17.17 & 17.2 min\\
3&6&2&12& 0.800 & 17.70 & 24.7 min\\
\midrule
3&1&2&10 & 0.798 & 17.17 & 37.6 min\\
3&2&2&10 & 0.798 & 17.17 & 25.6 min\\
3&3&2&10 & 0.798 & 17.17 & 24.5 min\\
3&4&2&10 & 0.798 & 17.17 & 20.3 min\\
3&5&2&10 & 0.798 & 17.17 & 18.5 min\\
3&6&2&10 & 0.798 & 17.17 & 17.2 min\\
\midrule
2&6&2&10 & 0.805 & 17.03 & 15.2 min\\
3&6&2&10 & 0.798 & 17.17 & 17.2 min\\
4&6&2&10 & 0.798 & 16.70 & 20.3 min\\
5&6&2&10 & 0.797 & 16.43 & 21.5 min\\
6&6&2&10 & 0.790 & 16.50 & 22.8 min\\
7&6&2&10 & 0.799 & 16.37 & 25.1 min\\
8&6&2&10 & 0.797 & 16.12 & 27.7 min\\
\bottomrule
\end{tabular}%
}
\vspace{-3mm}
\caption{\scriptsize Ablation study over \method{} hyperparameters on the \textit{engine} CT dataset under extreme shot noise (\emph{P50}).}
\label{tab:ablation}
\vspace{-3em}
\end{wraptable}
We perform an ablation study on the \emph{engine} CT dataset at the lowest SNR level (P50), to assess the influence of patch size, stride, batch size, and number of regions on reconstruction quality and runtime. 
For this experiment, we first reconstruct the 3D volume from its sparse, noisy projections using the standard CGLS algorithm, and then fit \method{} to the resulting volume under different parameter settings.

Results are reported in \Cref{tab:ablation}.
Increasing the patch size generally improves reconstruction quality in low-SNR settings, at the cost of increased runtime. We expect that the optimal patch size is likely an increasing function of noise level, as larger patches can average out more noise.
Using a stride of 2 achieves a favorable trade-off between reconstruction fidelity and efficiency; slightly higher fidelity is possible with a stride of 1, but at a substantial computational price.
The batch size for parallel processing primarily affects runtime, with no impact on reconstruction accuracy.
Varying the number of regions per junction shows that using two or three regions provides the best balance of reconstruction quality and speed. 
It is possible that higher junction complexity could prove beneficial for particularly complex structures, or with additional steps of refinement optimization.
Our default configuration uses a patch size of 10, stride of 2, batch size of 6, and $M=3$ regions per junction.
The complete set of ablation results, including additional configurations for initialization and refinement iterations, is provided in \Cref{tab:ablation_appendix} and \Cref{fig:ct_p50_slice_ablation_results_appendix} in the appendix. 
Additional task-scale runtime and peak-memory measurements are provided in \Cref{sec:app_compute_tradeoff}.

\vspace{-0.5em}
\section{Discussion}
\label{sec:discussion}
\vspace{-0.5em}
Our experiments confirm that our proposed 3D Field of Junctions parameterization is an effective prior for 3D denoising and noisy volumetric inverse problems, capable of extracting sharp boundary structure in 3D even under extremely low SNR. We emphasize that \method{} is fully explicit, with all parameters endowed with physical structural interpretation. It requires neither training data nor neural networks, can run on a single GPU or parallelize across several, and has fairly stable performance across hyperparameter settings and noise levels. It robustly extracts 3D structure across different types of noise, and works well in both direct volume denoising and indirect regularization of low-SNR volumetric inverse problems, outperforming the evaluated classical, untrained neural, and no-clean-target learned alternatives.

Our results are subject to three main limitations. First, we did not tune the various \method{} hyperparameters (patch size, stride, number of regions $M$ per patch, or the smoothness parameters $\eta$ and $\delta$) separately for each task. While this ensures our results reflect expected performance on new tasks, stronger performance is likely possible with additional hyperparameter tuning, especially for different noise levels and dataset resolutions.
A second limitation is computational cost. For a moderate-size volume ($256^3$), fitting \method{} runs in a few minutes on an NVIDIA A6000 GPU. However, memory usage increases with both patch and volume size, necessitating chunked or parallel processing, which our implementation supports, for large volumes. We provide task-scale runtime and peak-memory measurements for our CT, cryo-ET, and point-cloud denoising experiments in \Cref{sec:app_compute_tradeoff}.
Finally, \method{} imposes a local piecewise-planar inductive bias: curved or smoothly varying structures are represented by locally planar pieces. This bias is useful for suppressing noise and preserving sharp interfaces, and is well matched to many man-made objects and membrane-dominated biological structures. However, for highly curved structures, performance depends on the patch size and the number of active regions per patch. The curved-tube study in \Cref{sec:app_compute_tradeoff} shows that the default $M=3$ setting preserves curved geometry when the FoJ patch size is local relative to curvature, while using all $M=8$ sectors better handles large patches or high curvature, at higher memory and runtime cost.

\clearpage
\bibliographystyle{splncs04}
\bibliography{main}

\clearpage
\setcounter{page}{1}

\section{2D FoJ Optimization}\label{sec:appendix_2dfojoptimization}
The optimization objective in 2D Field of Junctions \cite{verbin2021field} is:
{
\footnotesize
\begin{equation}
\begin{aligned}
\arg\min_{\Theta,C}\quad
&\sum_{i,j}\!\iint u_{\theta_i}^{(j)}(\boldsymbol p)\,\|I_i(\boldsymbol p)-c_i^{(j)}\|^2\,d\boldsymbol p
+ \lambda_B\sum_i\!\iint_{\Omega_i}\![B_i(\boldsymbol p)-\widehat{B}(\boldsymbol p)]^2\,d\boldsymbol p \\
&\quad + \lambda_C\sum_{i,j}\!\iint u_{\theta_i}^{(j)}(\boldsymbol p)\,\|c_i^{(j)}-\widehat{I}(\boldsymbol p)\|^2\,d\boldsymbol p .
\end{aligned}
\label{eq:foj_objective}
\end{equation}
}

\noindent Here, $I_i(\boldsymbol p)$ denotes the observed (noisy) image within patch~$i$ with domain $\Omega_i$, and $c_i^{(j)}$ is the constant color associated with wedge~$j$ inside junction $i$.  
The first term of \Cref{eq:foj_objective} enforces fidelity between each observed image patch and its piecewise-constant junction representation.  
The second term, weighted by~$\lambda_B \ge 0$, aligns boundary locations among neighboring patches through the local and global boundary maps $B_i(\boldsymbol p)$ and $\widehat{B}(\boldsymbol p)$, respectively.  
The local map $B_i(\boldsymbol p)$ equals~1 at pixels lying on the boundaries defined by~$\theta_i$, and~0 elsewhere, while the global map~$\widehat{B}(\boldsymbol p)$ is obtained by averaging all overlapping local boundary maps across patches.  
The third term, weighted by~$\lambda_C \ge 0$, enforces consistent regional appearance by comparing each patch’s color estimate with the global color field~$\widehat{I}(\boldsymbol p)$, which averages the reconstructed intensities of all patches covering the same pixel.  
For optimization, the binary wedge indicator functions $u_{\theta_i}^{(j)}(\boldsymbol p)$ and boundary maps $B_i(\boldsymbol p)$ are replaced by smooth, differentiable approximations to enable gradient-based refinement; these differentiable formulations are detailed in \Cref{sec:soft region representation}.

\section{Soft Region Representation in 2D FoJ}\label{sec:soft region representation}

As described in \Cref{sec:method}, our \method{} uses smooth indicator functions to define soft, differentiable region assignments and boundary maps. These are inspired by a similar differentiable formulation in 2D FoJ \cite{verbin2021field}, which we describe here.


For each 2D patch, a set of signed distance functions is defined to measure the perpendicular distance of any pixel $\boldsymbol{p}=(x,y)$ from the $k$th boundary line passing through the vertex $\boldsymbol{p}_i^{(0)}$ at orientation $\phi_i^{(k)}$:
\begin{equation}
d_k(\boldsymbol{p}) = -(x-x_i^{(0)})\sin\phi_i^{(k)} + (y-y_i^{(0)})\cos\phi_i^{(k)}.
\label{eq:distance}
\end{equation}
Similarly, for each pair of boundary lines $k$ and $l$, the signed distance function is:
\begin{equation}
d_{kl}(\boldsymbol{p}) =
\begin{cases}
\min\{d_k(\boldsymbol{p}),-d_l(\boldsymbol{p})\}, & \text{if } \phi_i^{(l)}-\phi_i^{(k)} < \pi,\\[4pt]
\max\{d_k(\boldsymbol{p}),-d_l(\boldsymbol{p})\}, & \text{otherwise.}
\end{cases}
\label{eq:dkl}
\end{equation}

To create a differentiable indicator of junction structure, the binary wedge boundaries are replaced by 
differentiable functions based on a regularized Heaviside operator:
\begin{equation}
H_\eta(d)=\frac12\!\left(1+\frac{2}{\pi}\arctan\!\frac{d}{\eta}\right),
\label{eq:heaviside}
\end{equation}
where $\eta>0$ controls the smoothness of the transition between adjacent regions.  
For $M=3$, the continuous region indicators are then given by
\begin{align}
u_{\theta_i}^{(1)}(\boldsymbol{p})&=1-H_\eta(d_{13}(\textbf{p})), \nonumber\\
u_{\theta_i}^{(2)}(\boldsymbol{p})&=H_\eta(d_{13}(p))\,[1-H_\eta(d_{12}(\boldsymbol{p}))],\\
u_{\theta_i}^{(3)}(\boldsymbol{p})&=H_\eta(d_{13}(\boldsymbol{p}))\,H_\eta(d_{12}(\boldsymbol{p})).\nonumber
\label{eq:softu}
\end{align}
Each function $u_{\theta_i}^{(j)}(\boldsymbol{p})\in[0,1]$ varies smoothly across the boundary and approaches 1 inside region~$j$.  
This relaxation ensures differentiability of the objective function in \Cref{eq:foj_objective} with respect to both the angular and positional parameters of each junction.

The binary boundary maps $B_i(\boldsymbol{p})$ and $\widehat{B}(\boldsymbol{p})$ in \Cref{eq:foj_objective} are likewise replaced by soft local and global boundary maps, defined using the gradients of the smooth indicator functions.
The soft boundary map for patch~$i$ with $M=3$ is defined as
\begin{equation}
B_i^{(\delta)}(\boldsymbol{p})=\pi\delta\,H'_\delta\!\bigl(\min\{|d_{12}(\boldsymbol{p})|,|d_{13}(\boldsymbol{p})|\}\bigr),
\label{eq:softB}
\end{equation}
where $H'_\delta$ denotes the derivative of $H_\delta$ with respect to~$d$, and $\delta$ controls the effective width of the boundary transition.  
This continuous version approximates the binary boundary map while providing smoother gradients for optimization.
The 2D FoJ paper \cite{verbin2021field} recommends using 
$\eta{=}0.01$ and $\delta{=}0.1$, as these values yield sharp yet numerically stable transitions.

\section{3D FoJ Optimization Algorithm}
A full step-by-step description of the 3D FoJ optimization procedure is provided in \Cref{alg:foj3d}.

\begin{algorithm}[t]
\caption{3D FoJ Optimization Procedure}
\begin{algorithmic}[1]
\label{alg:foj3d}
\Require Volumetric data divided into 3D patches $\{\boldsymbol V_i\}$, number of patches $N$, number of planes $L$ per junction (default $L=3$), number of regions $M$ per junction (default $M=3$), initialization iterations $N_{\text{init}}$, refinement iterations $N_{\text{refine}}$, step size $\lambda$
\Ensure Optimized junction parameters $\boldsymbol\gamma_i = (\boldsymbol \rho_i, \boldsymbol v_i^{(0)})$ and region intensities $c_i^{(\ell)}$

\State \textbf{Initialization:}
\For{$i = 1, \ldots, N$} 
    \State Initialize offset $\boldsymbol v_i^{(0)}$ at patch center
    \For{$t = 1, \ldots, N_{\text{init}}$}
        \For{$\ell = 1, \ldots, L$}
            \State $(\theta_i^{(\ell)}, \phi_i^{(\ell)}) \leftarrow \arg\min_{\theta, \phi} \mathcal{L}_{\text{data}}(\theta, \phi)$
        \EndFor
        \State $\boldsymbol v_i^{(0)} \leftarrow \arg\min_{\boldsymbol v}\, \mathcal{L}_{\text{data}}(\boldsymbol \rho_i, \boldsymbol v)$
    \EndFor
\EndFor

\vspace{3pt}
\State \textbf{Refinement:}
\For{$t = 1, \ldots, N_{\text{refine}}$}
    \State Replace binary indicators $u_{\boldsymbol \gamma_i}^{(j)}$ and $B_i$ with differentiable versions $u_{\boldsymbol \gamma_i, \eta}^{(j)}$, $B_i^{(\delta)}$
    \For{$i = 1, \ldots, N$}
        \State Update orientation and offset parameters via gradient-based optimization:
        \[
        \boldsymbol \gamma_i \leftarrow \boldsymbol \gamma_i - \lambda \nabla_{\boldsymbol\gamma_i} \mathcal{L}_{\text{3D-FoJ}}
        \]
        \State Update region intensities analytically:
        \[
       c_i^{(j)} =
        \frac{\iiint u_{\boldsymbol \gamma_i, \eta}^{(j)}(\boldsymbol{v})\,[V_i(\boldsymbol{v})+\lambda_C\,\widehat{V}(\boldsymbol{v})]\,d\boldsymbol{v}}
        {(1+\lambda_C)\iiint u_{\boldsymbol \gamma_i, \eta}^{(j)}(\boldsymbol{v})\,d\boldsymbol{v}}
        \]
    \EndFor
    \State Gradually increase $\lambda_B, \lambda_C$ in the loss function, to enforce boundary and color consistency
\EndFor
\end{algorithmic}
\end{algorithm}

\section{Additional Results}
In this section, we present extended qualitative and quantitative evaluations complementing the main results in the paper.

\subsection{Low-dose X-ray Computed Tomography (CT)}
\Cref{tab:combined_PSNR_MSSSIM_appendix} reports MS-SSIM measured on all $360^\circ$ unseen projections (one projection per degree) and 3D PSNR measured on reconstructed volumes for every dataset (\textit{engine}, \textit{foot}, \textit{jaw}, \textit{pepper}, and \textit{teapot}) and noise level (\textit{P50}, \textit{P100}, and \textit{P1000}). 

\begin{table}[t]
\centering
\scriptsize
\setlength{\tabcolsep}{2.2pt}
\renewcommand{\arraystretch}{0.92}

\begin{minipage}[t]{0.49\linewidth}
\centering
\resizebox{\linewidth}{!}{%
\begin{tabular}{llccc}
\toprule
\multicolumn{5}{c}{\textbf{2D MS-SSIM}\,$\uparrow$} \\
\midrule
\textbf{Dataset} & \textbf{Method} & \textbf{P50} & \textbf{P100} & \textbf{P1000} \\
\midrule
\multirow{5}{*}{\textit{pepper}}
& \method{} (ours) & \textbf{0.698} & 0.736 & \textbf{0.780} \\
& R$^2$-Gaussian & 0.593 & 0.593 & 0.593 \\
& NAF & 0.420 & 0.455 & 0.527 \\
& Filter2Noise & 0.665 & \textbf{0.745} & 0.745 \\
& 3D-TV & 0.665 & 0.679 & 0.733 \\
\cmidrule(lr){1-5}

\multirow{5}{*}{\textit{teapot}}
& \method{} (ours) & \textbf{0.757} & \textbf{0.811} & 0.882 \\
& R$^2$-Gaussian & 0.692 & 0.692 & 0.651 \\
& NAF & 0.491 & 0.555 & 0.697 \\
& Filter2Noise & 0.534 & 0.547 & 0.531 \\
& 3D-TV & 0.668 & 0.747 & \textbf{0.939} \\
\cmidrule(lr){1-5}

\multirow{5}{*}{\textit{jaw}}
& \method{} (ours) & \textbf{0.917} & \textbf{0.914} & \textbf{0.912} \\
& R$^2$-Gaussian & 0.350 & 0.350 & 0.350 \\
& NAF & 0.388 & 0.416 & 0.490 \\
& Filter2Noise & 0.857 & 0.848 & 0.832 \\
& 3D-TV & 0.883 & 0.899 & 0.904 \\
\cmidrule(lr){1-5}

\multirow{5}{*}{\textit{foot}}
& \method{} (ours) & \textbf{0.630} & \textbf{0.660} & 0.704 \\
& R$^2$-Gaussian & 0.608 & 0.608 & 0.608 \\
& NAF & 0.408 & 0.448 & 0.560 \\
& Filter2Noise & 0.595 & 0.622 & 0.684 \\
& 3D-TV & 0.571 & 0.596 & \textbf{0.706} \\
\cmidrule(lr){1-5}

\multirow{5}{*}{\textit{engine}}
& \method{} (ours) & \textbf{0.766} & \textbf{0.820} & \textbf{0.911} \\
& R$^2$-Gaussian & 0.629 & 0.642 & 0.653 \\
& NAF & 0.408 & 0.461 & 0.569 \\
& Filter2Noise & 0.661 & 0.681 & 0.681 \\
& 3D-TV & 0.727 & 0.772 & 0.863 \\
\bottomrule
\end{tabular}}
\end{minipage}
\hfill
\begin{minipage}[t]{0.49\linewidth}
\centering
\resizebox{\linewidth}{!}{%
\begin{tabular}{llccc}
\toprule
\multicolumn{5}{c}{\textbf{3D PSNR}\,$\uparrow$} \\
\midrule
\textbf{Dataset} & \textbf{Method} & \textbf{P50} & \textbf{P100} & \textbf{P1000} \\
\midrule
\multirow{5}{*}{\textit{pepper}}
& \method{} (ours) & 12.13 & 12.54 & 13.49 \\
& R$^2$-Gaussian & \textbf{13.48} & \textbf{13.48} & 13.48 \\
& NAF & 10.47 & 10.84 & 11.23 \\
& Filter2Noise & 6.93 & 13.42 & 13.42 \\
& 3D-TV & 8.65 & 10.92 & \textbf{13.86} \\
\cmidrule(lr){1-5}

\multirow{5}{*}{\textit{teapot}}
& \method{} (ours) & \textbf{20.45} & 20.69 & \textbf{21.18} \\
& R$^2$-Gaussian & 18.55 & 18.55 & 17.50 \\
& NAF & 18.28 & 19.06 & 20.67 \\
& Filter2Noise & 16.38 & 18.20 & 11.15 \\
& 3D-TV & 19.90 & \textbf{20.72} & 21.16 \\
\cmidrule(lr){1-5}

\multirow{5}{*}{\textit{jaw}}
& \method{} (ours) & \textbf{18.74} & 18.81 & \textbf{19.00} \\
& R$^2$-Gaussian & 12.55 & 12.55 & 12.55 \\
& NAF & 15.71 & 15.89 & 17.35 \\
& Filter2Noise & 12.93 & 14.19 & 15.70 \\
& 3D-TV & 18.64 & \textbf{19.09} & \textbf{19.00} \\
\cmidrule(lr){1-5}

\multirow{5}{*}{\textit{foot}}
& \method{} (ours) & \textbf{15.60} & 15.79 & 16.15 \\
& R$^2$-Gaussian & 15.26 & 15.26 & 15.26 \\
& NAF & 14.94 & 15.32 & 15.75 \\
& Filter2Noise & 11.87 & 13.74 & \textbf{16.80} \\
& 3D-TV & 14.40 & \textbf{15.84} & 16.71 \\
\cmidrule(lr){1-5}

\multirow{5}{*}{\textit{engine}}
& \method{} (ours) & 12.49 & 12.86 & 13.77 \\
& R$^2$-Gaussian & \textbf{16.61} & \textbf{16.61} & \textbf{16.63} \\
& NAF & 11.46 & 12.22 & 13.28 \\
& Filter2Noise & 9.93 & 16.11 & 16.11 \\
& 3D-TV & 7.37 & 9.96 & 14.08 \\
\bottomrule
\end{tabular}}
\end{minipage}

\caption{Per-dataset CT quantitative comparison. MS-SSIM is measured on 2D projections across the full $360^\circ$ range, and 3D PSNR is measured on reconstructed 3D volumes. The best value for each dataset and noise level is bolded.}
\label{tab:combined_PSNR_MSSSIM_appendix}
\vspace{-1em}
\end{table}

\begin{figure}[h]
  \centering
    \includegraphics[width=0.9\linewidth]{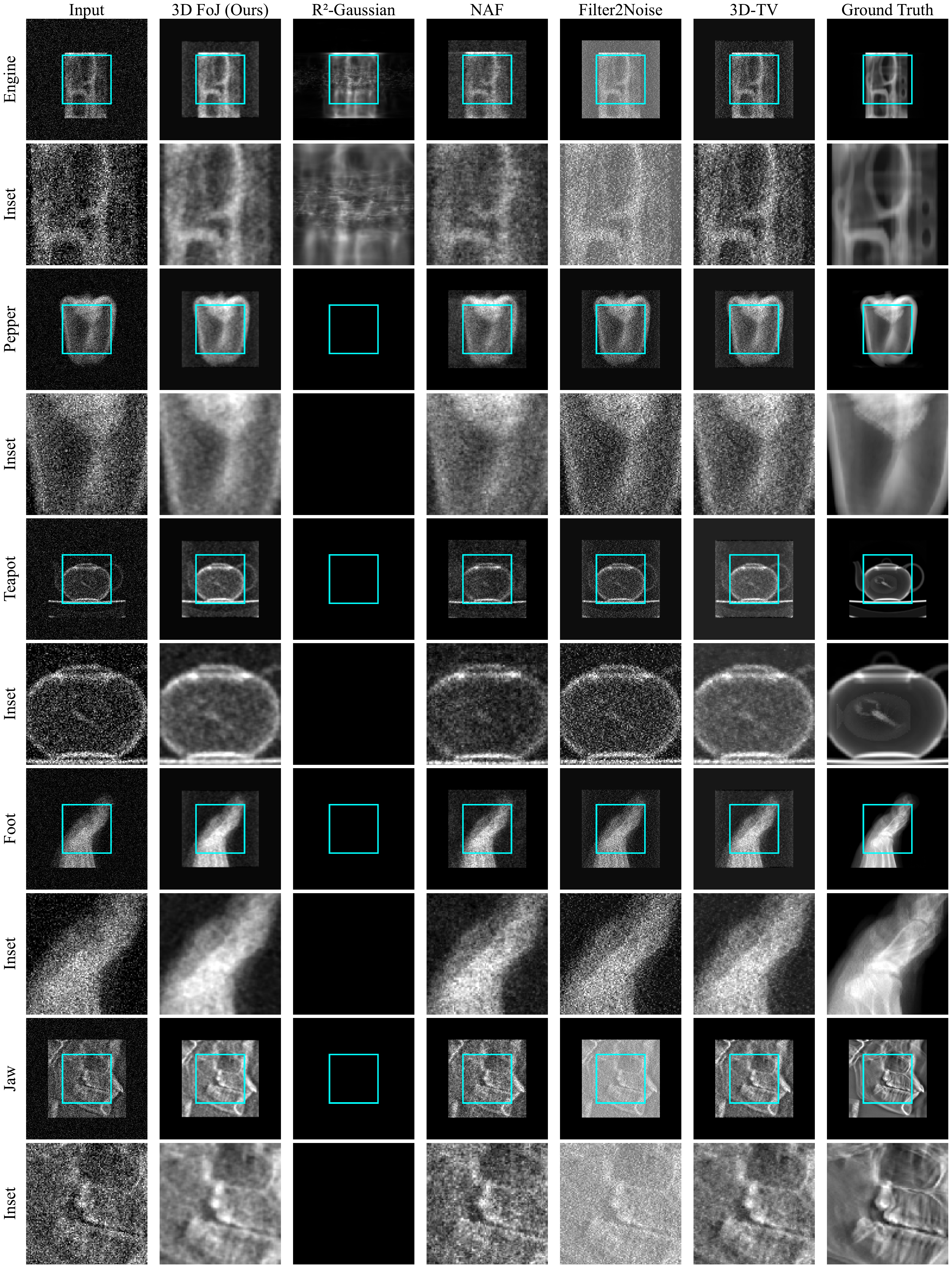}
    \caption{Comparison of unseen projection views synthesized from reconstructed 3D volumes for all datasets under low-SNR (\textit{P50}) conditions.}
  \vspace{-1.5em}
  \label{fig:ct_p50_results_appendix}
\end{figure}

\begin{figure}[h]
  \centering
    \includegraphics[width=0.9\linewidth]{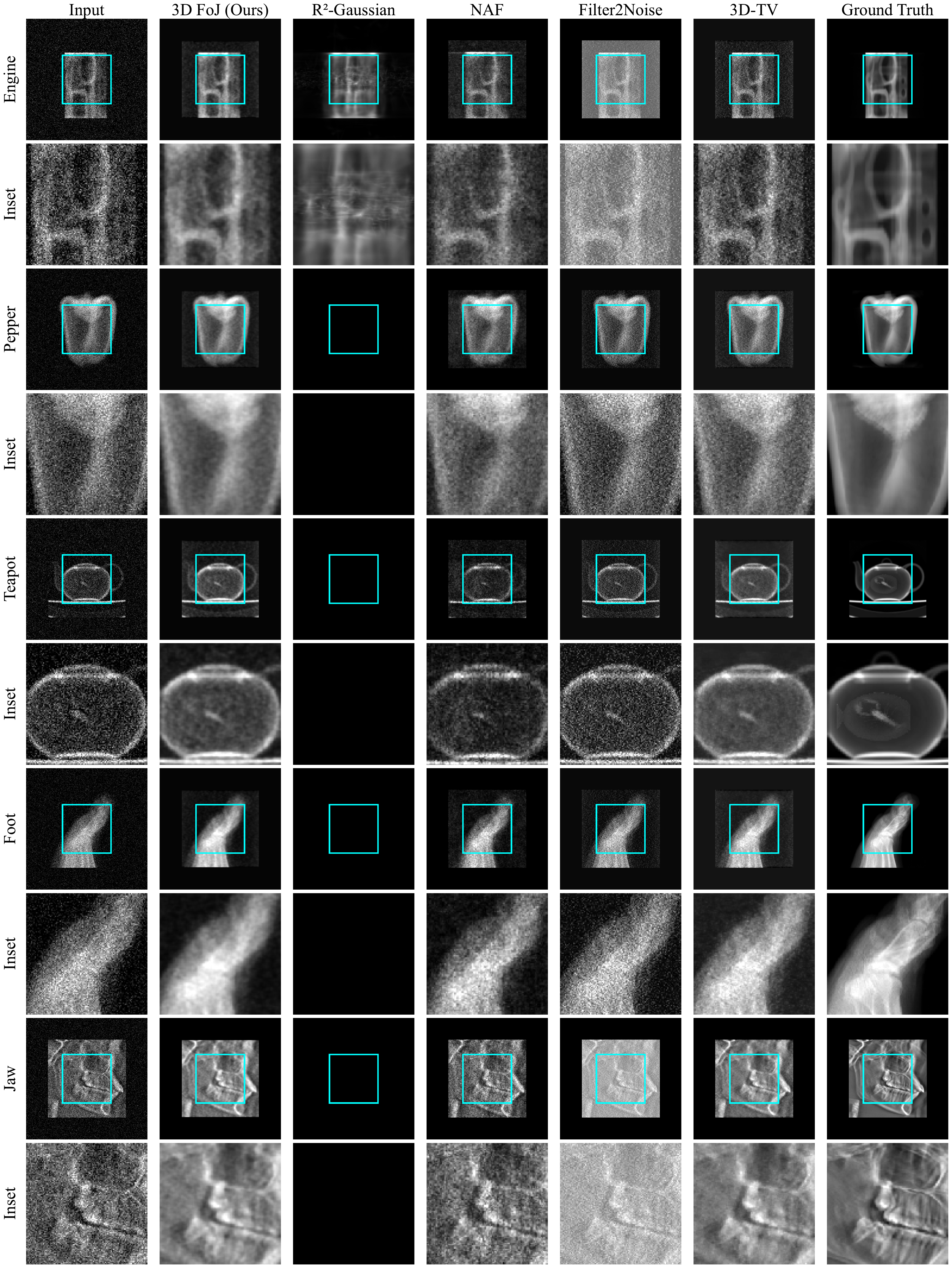}
    \caption{Comparison of unseen projection views synthesized from reconstructed 3D volumes for all datasets under medium-SNR (\textit{P100}) conditions.}
  \vspace{-1.5em}
  \label{fig:ct_p100_results_appendix}
\end{figure}

\begin{figure}[h]
  \centering
    \includegraphics[width=0.85\linewidth]{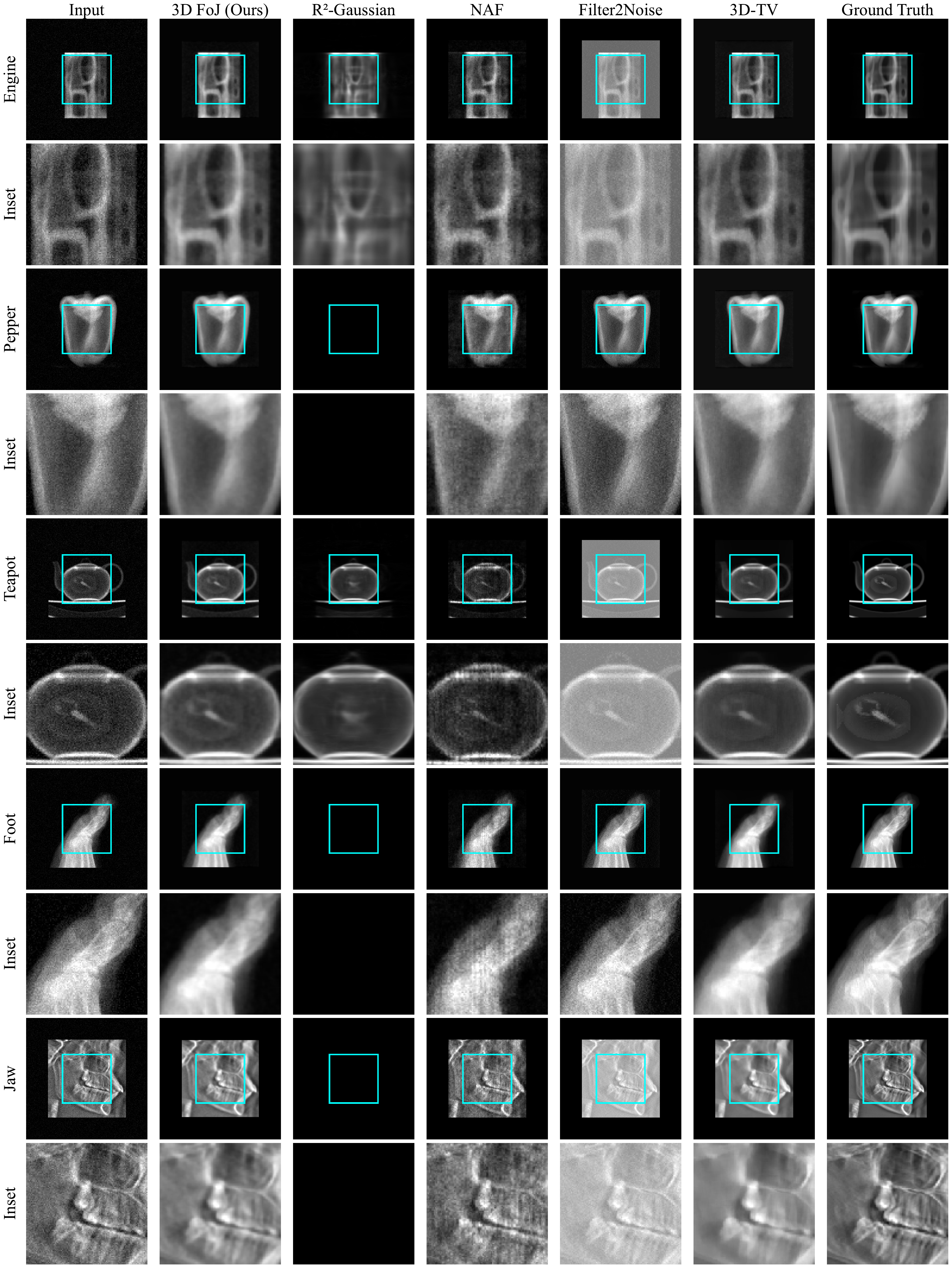}
    \caption{Comparison of unseen projection views synthesized from reconstructed 3D volumes for all datasets under high-SNR (\textit{P1000}) conditions.}
  \vspace{-1.5em}
  \label{fig:ct_p1000_results_appendix}
\end{figure}

The quantitative results in \Cref{tab:combined_PSNR_MSSSIM_appendix} show that \method{} achieves the best projection MS-SSIM in most dataset/noise settings, especially at the lowest-SNR condition (\textit{P50}), while 3D PSNR is more mixed: R$^2$-Gaussian, 3D-TV, and Filter2Noise obtain the highest PSNR for several dataset/noise pairs, particularly at higher SNR. 
We therefore interpret the scalar metrics together with qualitative projection and slice comparisons. 
\Cref{fig:ct_p50_results_appendix,fig:ct_p100_results_appendix,fig:ct_p1000_results_appendix} visualize unseen projection views synthesized from the reconstructed 3D volumes at noise levels \textit{P50}, \textit{P100}, and \textit{P1000}, respectively. 
Across these visual comparisons, \method{} recovers sharper boundaries, more accurate object contours, and fewer noise-induced artifacts, especially under low-SNR conditions. 
In contrast, R$^2$-Gaussian and Filter2Noise often lose fine structural details, while 3D-TV tends to over-smooth textures, resulting in blurred edges and degraded object shape.

\begin{figure}[h]
  \centering
    \includegraphics[width=0.6\linewidth]{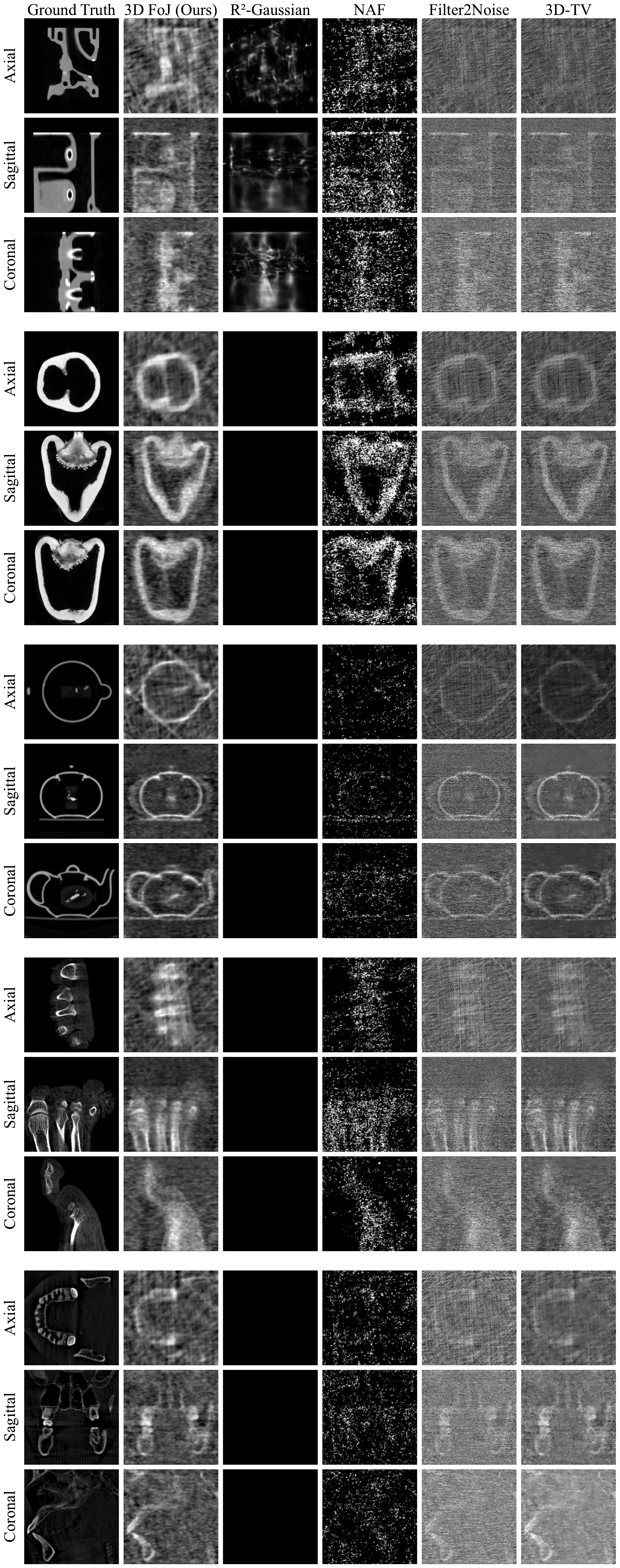}
    \caption{Comparison of slice views of reconstructed 3D volumes for all datasets under low-SNR (\textit{P50}) conditions.}
  \vspace{-1.5em}
  \label{fig:ct_p50_slice_results_appendix}
\end{figure}

\begin{figure}[h]
  \centering
    \includegraphics[width=0.6\linewidth]{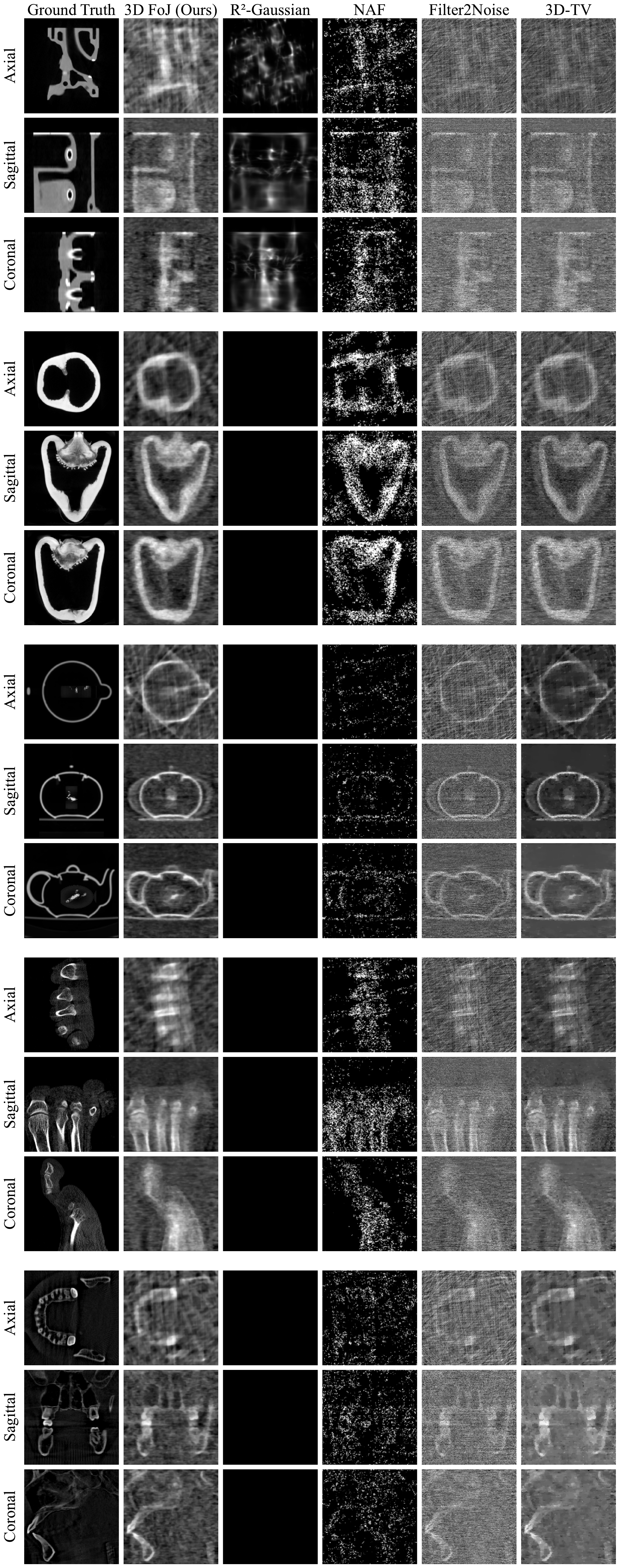}
    \caption{Comparison of slice views of reconstructed 3D volumes for all datasets under medium-SNR (\textit{P100}) conditions.}
  \vspace{-1.5em}
  \label{fig:ct_p100_slice_results_appendix}
\end{figure}

\begin{figure}[h]
  \centering
    \includegraphics[width=0.6\linewidth]{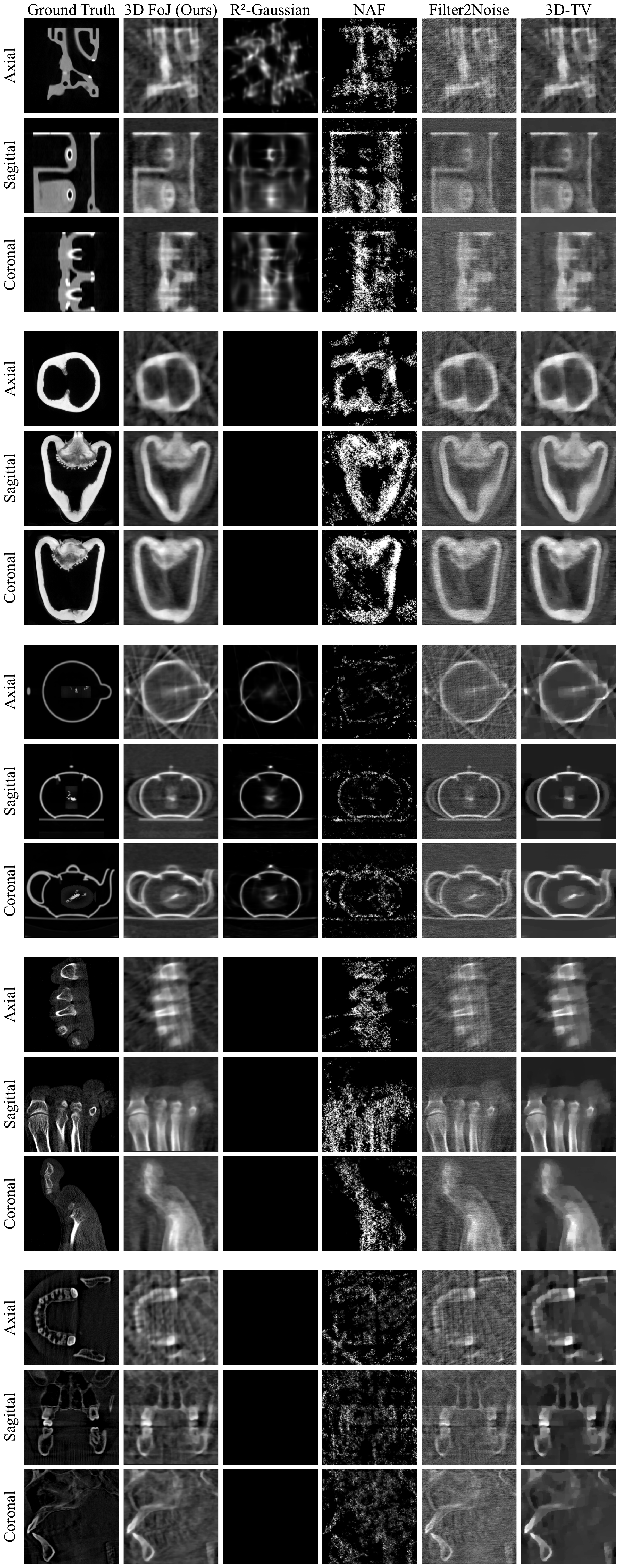}
    \caption{Comparison of slice views of reconstructed 3D volumes for all datasets under high-SNR (\textit{P1000}) conditions.}
  \vspace{-1.5em}
  \label{fig:ct_p1000_slice_results_appendix}
\end{figure}

\Cref{fig:ct_p50_slice_results_appendix,fig:ct_p100_slice_results_appendix,fig:ct_p1000_slice_results_appendix} present axial, sagittal, and coronal slices of reconstructed 3D volumes across datasets at noise levels \textit{P50}, \textit{P100}, and \textit{P1000}, respectively. 
These slice views show that \method{} preserves boundaries, interfaces, and material transitions more consistently than other methods, while R$^2$-Gaussian often overfits to the limited projection views and fails to recover coherent 3D structure. 
TV-based reconstructions exhibit blocky artifacts and detail loss, while Filter2Noise reduces noise but can remove fine boundaries. 
Together, the projection and slice comparisons show that \method{} better preserves internal geometry under highly degraded sparse-view and low-dose CT conditions, even when PSNR alone does not rank it first in every setting.

\subsection{Cryogenic Electron Tomography (cryo-ET) }

\begin{figure}[h]
  \centering
    \includegraphics[width=0.6\linewidth]{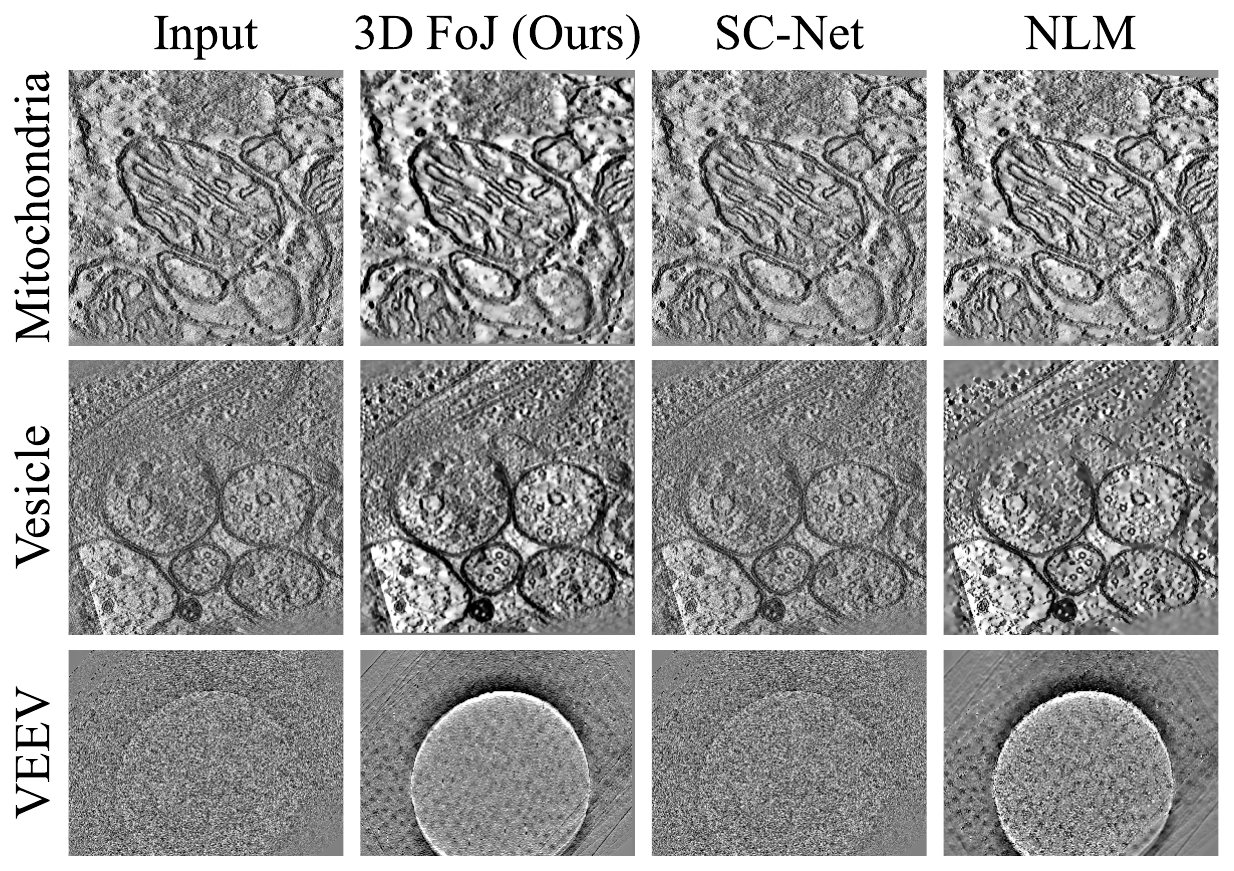}
    \caption{Additional visual denoising results for the real cryo-ET \emph{Mitochondria, Vesicle}, and {VEEV} volumes, sliced in the $y$ dimension. \method{} enhances membrane contrast and resolves fine structural details while reducing noise, compared with the input, SC-Net~\cite{cryoet2021}, and nonlocal means (NLM)~\cite{nonlocalmeansforCT2017}.}
  \vspace{-1.5em}
  \label{fig:cryoet_results_appendix}
\end{figure}

We provide additional qualitative denoising results for three real cryo-ET datasets, \emph{mitochondria}, \emph{vesicle}, and \emph{VEEV}, in \Cref{fig:cryoet_results_appendix}. As in the main paper, all methods operate directly on the noisy 3D reconstructions, which contain both strong Poisson noise from low electron dose and missing-wedge artifacts from limited tilt angles. 

Across all three datasets, \method{} produces the best trade-off between noise removal, contrast enhancement, and preservation of fine structures. In the \emph{mitochondria} volume, \method{} sharpens inner and outer mitochondrial membranes, whereas SC-Net~\cite{cryoet2021} and NLM~\cite{nonlocalmeansforCT2017} over-smooth these thin membranes and suppress low-contrast internal detail. For the \emph{vesicle} volume, \method{} enhances vesicle boundaries and resolves small sub-vesicular structures while suppressing the granular background noise that remains in the input and NLM outputs. For the \emph{VEEV} dataset, \method{} recovers a clear, smooth viral boundary while simultaneously enhancing the thin, elongated groove-like structure in the upper-left region of the slice. This feature is faint and partially obscured in the noisy input and NLM outputs, and SC-Net tends to over-smooth it together with the surrounding membrane. In contrast, \method{} preserves the narrow linear indentation with high contrast, revealing its full shape. This example highlights the ability of \method{} to recover subtle, low-contrast morphological details. Together with the \emph{centriole} results shown in the main paper, these examples indicate that \method{} can reliably denoise real cryo-ET volumes while preserving delicate biological morphology in the absence of ground-truth supervision.

\subsection{Point Cloud Denoising}
\label{point_cloud_denoising_appendix}
We show the behavior of each method across 28 point cloud datasets under two noise settings: outlier noise and spread noise. The comparison metric is Chamfer-L2 error; results are reported in \Cref{tab:outlier_ChamferL2_2x2_levels,tab:spread_ChamferL2_2x2_levels}.

For voxel-based methods \method{}, nonlocal means (NLM)~\cite{nonlocalmeansforCT2017}, and 3D TV~\cite{TV}, we use an identical conversion protocol across methods. 
We define a shared $256^3$ voxel grid in the benchmark reference coordinate frame, anchored to the ground-truth coordinate bounds with 10-voxel padding. 
This anchoring is used only to define a common evaluation grid and is shared across all voxel-based methods. 
Input points are assigned to their nearest voxel centers to form the voxelized input. 
After denoising, voxel outputs are converted back to point clouds by selecting the top $K=100{,}000$ voxels by output intensity and mapping their centers back to world coordinates, with fixed-seed within-voxel jitter.

In the outlier-noise setting, \method{} achieves the lowest Chamfer-L2 error on every dataset and at every outlier ratio (10\%, 30\%, 60\%, 90\%). While PointCVaR~\cite{li2024pointcvar} and PointCleanNet~\cite{rakotosaona2020pointcleannet} show competitive performance at low outlier ratios, their fidelity degrades substantially as the proportion of outliers grows. NLM~\cite{nonlocalmeansforCT2017} exhibits similar degradation under high noise. In contrast, \method{} preserves accurate shape even at the highest outlier levels.

We can observe a similar trend in the spread-noise setting. Across all four noise levels (40k, 100k, 200k, 500k), \method{} maintains the lowest or near-lowest Chamfer-L2 error across different shapes. PointCVaR performs competitively at the 100k noise level, but its errors rise at higher noise levels. PointCleanNet and NLM also display a consistent increase in error as noise increases. In comparison, \method{} exhibits significantly lower error even under the highest noise, showing robustness across all spread-noise conditions.

Overall, these quantitative results demonstrate that \method{} provides the most stable point cloud denoising performance across both outlier and spread noise models and all noise levels, with consistently lower Chamfer-L2 errors than the competing methods.

\begin{table*}[h]
\centering
\resizebox{0.85\textwidth}{!}{%
\begin{tabular}{l|cccc|cccc}
\toprule
 & \multicolumn{4}{c}{Noise Level 10\%} & \multicolumn{4}{c}{Noise Level 30\%} \\
Dataset & PointCVaR & PointCleanNet & NLM & \method{} & PointCVaR & PointCleanNet & NLM & \method{} \\
\midrule
Cup33 & 0.0014 & 0.0020 & 0.0018 & 0.0001 & 0.0050 & 0.0060 & 0.0058 & 0.0001 \\
LegoLeg & 0.0016 & 0.0017 & 0.0017 & 0.0001 & 0.0051 & 0.0052 & 0.0051 & 0.0001 \\
Liberty & 0.0296 & 0.0322 & 0.1299 & 0.0015 & 0.0925 & 0.0964 & 0.1466 & 0.0016 \\
LibertyBase & 0.0024 & 0.0026 & 0.0020 & 0.0002 & 0.0074 & 0.0077 & 0.0062 & 0.0003 \\
armadillo & 17.4917 & 17.4023 & 41.1563 & 1.0438 & 53.2626 & 53.0428 & 60.0359 & 1.0914 \\
box\_groove & 0.0468 & 0.0553 & 0.0159 & 0.0047 & 0.1531 & 0.1650 & 0.0760 & 0.0120 \\
box\_push & 0.0730 & 0.0797 & 0.0729 & 0.0055 & 0.2298 & 0.2396 & 0.2363 & 0.0062 \\
boxunion2 & 0.0032 & 0.0047 & 0.0045 & 0.0003 & 0.0116 & 0.0140 & 0.0135 & 0.0003 \\
boxunion\_uniform & 0.0023 & 0.0032 & 0.0009 & 0.0003 & 0.0081 & 0.0095 & 0.0060 & 0.0003 \\
bunny & 0.0067 & 0.0080 & 0.0079 & 0.0004 & 0.0223 & 0.0242 & 0.0241 & 0.0005 \\
column & 3.5018 & 3.4981 & 8.8556 & 0.2346 & 10.2025 & 10.6478 & 13.0056 & 0.1701 \\
column\_head & 0.2254 & 0.2231 & 0.1915 & 0.0463 & 0.6704 & 0.6673 & 0.5965 & 0.0522 \\
cube\_uniform & 0.0006 & 0.0009 & 0.0002 & 0.0001 & 0.0023 & 0.0027 & 0.0009 & 0.0001 \\
cylinder & 0.0027 & 0.0038 & 0.0023 & 0.0002 & 0.0097 & 0.0114 & 0.0072 & 0.0003 \\
dragon & 0.0000 & 0.0000 & 0.0000 & 0.0000 & 0.0001 & 0.0001 & 0.0001 & 0.0000 \\
dragon\_xyzrgb & 30.9469 & 30.7425 & 82.1093 & 1.7697 & 92.1006 & 91.3520 & 118.2381 & 1.8064 \\
fandisk & 0.0194 & 0.0212 & 0.0208 & 0.0011 & 0.0604 & 0.0631 & 0.0664 & 0.0011 \\
flower & 0.8129 & 0.9397 & 1.0486 & 0.0504 & 2.6427 & 2.8260 & 3.3228 & 0.0537 \\
galera & 0.0163 & 0.0166 & 0.0169 & 0.0011 & 0.0497 & 0.0500 & 0.0518 & 0.0011 \\
happy & 0.0000 & 0.0000 & 0.0001 & 0.0000 & 0.0001 & 0.0001 & 0.0001 & 0.0000 \\
icosahedron & 0.0019 & 0.0027 & 0.0023 & 0.0001 & 0.0070 & 0.0082 & 0.0074 & 0.0002 \\
netsuke & 2.8787 & 2.8236 & 2.9063 & 0.1953 & 8.6102 & 8.5556 & 9.1302 & 0.1844 \\
pipe & 0.0150 & 0.0168 & 0.0173 & 0.0010 & 0.0478 & 0.0506 & 0.0513 & 0.0010 \\
pipe\_curve & 0.0168 & 0.0184 & 0.0197 & 0.0012 & 0.0527 & 0.0551 & 0.0576 & 0.0012 \\
star\_halfsmooth & 0.0074 & 0.0075 & 0.0078 & 0.0004 & 0.0225 & 0.0228 & 0.0261 & 0.0004 \\
star\_smooth & 0.0037 & 0.0046 & 0.0048 & 0.0003 & 0.0123 & 0.0139 & 0.0152 & 0.0003 \\
tortuga & 0.0031 & 0.0033 & 0.0060 & 0.0002 & 0.0097 & 0.0101 & 0.0126 & 0.0002 \\
yoda & 4.0502 & 4.9398 & 3.8923 & 0.8875 & 14.6877 & 14.6165 & 13.9355 & 1.0102 \\
\midrule
 & \multicolumn{4}{c}{Noise Level 60\%} & \multicolumn{4}{c}{Noise Level 90\%} \\
Dataset & PointCVaR & PointCleanNet & NLM & \method{} & PointCVaR & PointCleanNet & NLM & \method{} \\
\midrule
Cup33 & 0.0109 & 0.0120 & 0.0110 & 0.0002 & 0.0170 & 0.0181 & 0.0163 & 0.0030 \\
LegoLeg & 0.0103 & 0.0104 & 0.0095 & 0.0001 & 0.0155 & 0.0157 & 0.0137 & 0.0014 \\
Liberty & 0.1877 & 0.1930 & 0.2124 & 0.0032 & 0.2866 & 0.2906 & 0.2717 & 0.0205 \\
LibertyBase & 0.0152 & 0.0157 & 0.0118 & 0.0006 & 0.0232 & 0.0237 & 0.0175 & 0.0100 \\
armadillo & 106.0517 & 106.0869 & 103.7139 & 1.2459 & 160.9545 & 160.5282 & 145.6340 & 8.4291 \\
box\_groove & 0.3154 & 0.3303 & 0.1476 & 0.0178 & 0.4852 & 0.4989 & 0.2250 & 0.1640 \\
box\_push & 0.4662 & 0.4777 & 0.4372 & 0.0105 & 0.7112 & 0.7199 & 0.6439 & 0.1695 \\
boxunion2 & 0.0257 & 0.0281 & 0.0250 & 0.0004 & 0.0398 & 0.0424 & 0.0364 & 0.0063 \\
boxunion\_uniform & 0.0174 & 0.0190 & 0.0150 & 0.0004 & 0.0270 & 0.0285 & 0.0221 & 0.0071 \\
bunny & 0.0459 & 0.0479 & 0.0432 & 0.0007 & 0.0704 & 0.0722 & 0.0617 & 0.0108 \\
column & 20.8850 & 21.1058 & 21.7367 & 0.2624 & 31.8108 & 31.6407 & 28.8462 & 2.6576 \\
column\_head & 1.3346 & 1.3331 & 1.1293 & 0.0692 & 2.0186 & 2.0092 & 1.6877 & 0.4316 \\
cube\_uniform & 0.0049 & 0.0055 & 0.0023 & 0.0003 & 0.0077 & 0.0082 & 0.0036 & 0.0022 \\
cylinder & 0.0206 & 0.0227 & 0.0137 & 0.0009 & 0.0321 & 0.0341 & 0.0207 & 0.0104 \\
dragon & 0.0001 & 0.0002 & 0.0002 & 0.0000 & 0.0002 & 0.0002 & 0.0002 & 0.0000 \\
dragon\_xyzrgb & 181.3145 & 180.7101 & 197.2864 & 2.5481 & 273.3175 & 271.9570 & 263.5173 & 29.7845 \\
fandisk & 0.1228 & 0.1260 & 0.1177 & 0.0012 & 0.1878 & 0.1910 & 0.1690 & 0.0175 \\
flower & 5.4071 & 5.6100 & 5.7536 & 0.0639 & 8.2968 & 8.4817 & 7.9777 & 0.5627 \\
galera & 0.0997 & 0.1002 & 0.0954 & 0.0013 & 0.1503 & 0.1502 & 0.1356 & 0.0149 \\
happy & 0.0001 & 0.0002 & 0.0002 & 0.0000 & 0.0002 & 0.0003 & 0.0002 & 0.0000 \\
icosahedron & 0.0151 & 0.0166 & 0.0138 & 0.0003 & 0.0237 & 0.0250 & 0.0203 & 0.0074 \\
netsuke & 17.2499 & 17.2595 & 16.4935 & 0.2323 & 26.0029 & 25.9334 & 23.4266 & 2.1315 \\
pipe & 0.0980 & 0.1016 & 0.0927 & 0.0015 & 0.1492 & 0.1527 & 0.1309 & 0.0150 \\
pipe\_curve & 0.1078 & 0.1109 & 0.1053 & 0.0017 & 0.1628 & 0.1659 & 0.1466 & 0.0233 \\
star\_halfsmooth & 0.0451 & 0.0456 & 0.0466 & 0.0004 & 0.0682 & 0.0685 & 0.0655 & 0.0068 \\
star\_smooth & 0.0274 & 0.0277 & 0.0276 & 0.0003 & 0.0415 & 0.0417 & 0.0393 & 0.0032 \\
tortuga & 0.0198 & 0.0201 & 0.0213 & 0.0002 & 0.0299 & 0.0302 & 0.0287 & 0.0012 \\
yoda & 28.1077 & 29.6142 & 27.8578 & 1.7628 & 44.9544 & 44.8306 & 40.5248 & 4.3073 \\
\bottomrule
\end{tabular}}
\caption{ChamferL2 comparison under the outlier noise settings \{10\%, 30\%, 60\%, 90\%\}.
\method{} achieves the lowest error across all datasets and remains stable even as the outlier ratio increases.
Other methods show rapidly increasing errors at higher noise levels, especially on complex shapes such as armadillo, dragon, and flower, while \method{} preserves geometry with only a small change in error.
Overall, \method{} provides the most reliable reconstruction quality across all outlier noise levels.}
\label{tab:outlier_ChamferL2_2x2_levels}
\end{table*}

\begin{table*}[h]
\centering
\resizebox{0.85\textwidth}{!}{%
\begin{tabular}{l|cccc|cccc}
\toprule
 & \multicolumn{4}{c}{Noise Level 40000} & \multicolumn{4}{c}{Noise Level 100000} \\
Dataset & PointCVaR & PointCleanNet & NLM & \method{} & PointCVaR & PointCleanNet & NLM & \method{} \\
\midrule
Cup33 & 30.9605 & 0.0248 & 0.0293 & 0.0001 & 1.7726 & 0.0435 & 0.0483 & 0.0002 \\
LegoLeg & 30.3808 & 0.0304 & 0.0372 & 0.0001 & 0.0000 & 0.0532 & 0.0604 & 0.0001 \\
Liberty & 40.4311 & 1.2590 & 2.2693 & 0.0015 & 0.0117 & 2.2111 & 3.1473 & 0.0014 \\
LibertyBase & 28.3421 & 0.0100 & 0.0113 & 0.0002 & 0.0878 & 0.0175 & 0.0188 & 0.0002 \\
armadillo & 525.1897 & 377.7829 & 503.1765 & 1.0885 & 0.0000 & 660.6212 & 796.6649 & 1.1087 \\
box\_groove & 21.3624 & 0.2838 & 0.3158 & 0.0062 & 0.0000 & 0.4977 & 0.5290 & 0.0056 \\
box\_push & 28.3982 & 0.3628 & 0.4220 & 0.0058 & 0.0000 & 0.6338 & 0.6988 & 0.0057 \\
boxunion2 & 30.6402 & 0.0547 & 0.0663 & 0.0003 & 1.5264 & 0.0962 & 0.1092 & 0.0003 \\
boxunion\_uniform & 29.5142 & 0.0207 & 0.0239 & 0.0003 & 1.4309 & 0.0364 & 0.0397 & 0.0003 \\
bunny & 30.9756 & 0.1229 & 0.1526 & 0.0005 & 1.3982 & 0.2131 & 0.2464 & 0.0004 \\
column & 514.2000 & 141.3988 & 196.5365 & 0.1645 & 0.0000 & 247.0045 & 326.9531 & 0.1643 \\
column\_head & 30.4894 & 1.9100 & 2.0876 & 0.0410 & 0.0000 & 3.3705 & 3.5892 & 0.0439 \\
cube\_uniform & 28.9115 & 0.0056 & 0.0062 & 0.0001 & 1.9634 & 0.0098 & 0.0104 & 0.0001 \\
cylinder & 28.4080 & 0.0212 & 0.0243 & 0.0002 & 1.1429 & 0.0376 & 0.0408 & 0.0002 \\
dragon & 30.1898 & 0.0008 & 0.0011 & 0.0000 & 0.7854 & 0.0014 & 0.0017 & 0.0000 \\
dragon\_xyzrgb & 2094.9787 & 972.3282 & 1438.1343 & 1.6082 & 0.0000 & 1687.3487 & 2253.5707 & 1.7542 \\
fandisk & 32.0970 & 0.4419 & 0.5664 & 0.0011 & 0.0000 & 0.7744 & 0.9097 & 0.0011 \\
flower & 85.8532 & 20.9525 & 27.6863 & 0.0465 & 0.0000 & 36.3354 & 43.6728 & 0.0510 \\
galera & 33.0891 & 0.3065 & 0.3777 & 0.0011 & 0.0000 & 0.5393 & 0.6196 & 0.0012 \\
happy & 30.2999 & 0.0010 & 0.0014 & 0.0000 & 0.7180 & 0.0018 & 0.0022 & 0.0000 \\
icosahedron & 30.4123 & 0.0186 & 0.0219 & 0.0001 & 1.3710 & 0.0325 & 0.0361 & 0.0002 \\
netsuke & 137.3421 & 68.2139 & 85.3554 & 0.1837 & 0.0000 & 119.0337 & 137.7499 & 0.2154 \\
pipe & 35.1910 & 0.5018 & 0.6291 & 0.0010 & 0.0000 & 0.8755 & 1.0136 & 0.0010 \\
pipe\_curve & 34.7353 & 0.4732 & 0.5875 & 0.0012 & 0.0000 & 0.8271 & 0.9479 & 0.0014 \\
star\_halfsmooth & 27.1354 & 0.0694 & 0.0902 & 0.0004 & 0.0000 & 0.1215 & 0.1447 & 0.0004 \\
star\_smooth & 27.9685 & 0.0360 & 0.0446 & 0.0003 & 0.0016 & 0.0626 & 0.0718 & 0.0003 \\
tortuga & 31.6051 & 0.0721 & 0.1000 & 0.0002 & 0.0000 & 0.1260 & 0.1586 & 0.0002 \\
yoda & 294.4679 & 158.4453 & 178.6378 & 1.2085 & 0.0000 & 277.1665 & 296.8689 & 1.3129 \\
\midrule
 & \multicolumn{4}{c}{Noise Level 200000} & \multicolumn{4}{c}{Noise Level 500000} \\
Dataset & PointCVaR & PointCleanNet & NLM & \method{} & PointCVaR & PointCleanNet & NLM & \method{} \\
\midrule
Cup33 & 51.2295 & 0.0581 & 0.0616 & 0.0002 & 101.6741 & 0.0730 & 0.0757 & 0.0003 \\
LegoLeg & 50.1370 & 0.0711 & 0.0773 & 0.0001 & 102.0256 & 0.0889 & 0.0828 & 0.0001 \\
Liberty & 66.7135 & 2.9550 & 3.6719 & 0.0015 & 135.8302 & 3.6894 & 3.6443 & 0.0017 \\
LibertyBase & 46.7883 & 0.0232 & 0.0244 & 0.0002 & 94.8299 & 0.0289 & 0.0039 & 0.0004 \\
armadillo & 869.0288 & 879.8691 & 987.8753 & 1.1186 & 1737.7334 & 1098.1857 & 438.2211 & 1.3077 \\
box\_groove & 35.8121 & 0.6634 & 0.6907 & 0.0057 & 71.6800 & 0.8265 & 0.7507 & 0.0076 \\
box\_push & 47.4676 & 0.8404 & 0.8982 & 0.0064 & 96.6765 & 1.0576 & 0.1881 & 0.0074 \\
boxunion2 & 50.5676 & 0.1287 & 0.1395 & 0.0003 & 101.2901 & 0.1606 & 0.0429 & 0.0004 \\
boxunion\_uniform & 48.7418 & 0.0486 & 0.0519 & 0.0003 & 97.0436 & 0.0606 & 0.0106 & 0.0004 \\
bunny & 51.0638 & 0.2854 & 0.3128 & 0.0005 & 102.1750 & 0.3560 & 0.0628 & 0.0006 \\
column & 844.1674 & 329.7473 & 392.9127 & 0.1895 & 1704.4415 & 411.7895 & 407.2683 & 0.1811 \\
column\_head & 50.7623 & 4.4689 & 4.6424 & 0.0436 & 101.7552 & 5.5979 & 0.7142 & 0.0422 \\
cube\_uniform & 47.7497 & 0.0132 & 0.0104 & 0.0001 & 95.0206 & 0.0164 & 0.0155 & 0.0002 \\
cylinder & 47.4803 & 0.0497 & 0.0514 & 0.0003 & 94.9279 & 0.0623 & 0.0570 & 0.0004 \\
dragon & 49.6818 & 0.0019 & 0.0021 & 0.0000 & 100.2598 & 0.0023 & 0.0009 & 0.0000 \\
dragon\_xyzrgb & 3503.9741 & 2259.3765 & 2696.7554 & 1.9382 & 7024.6872 & 2825.4123 & 1740.8934 & 2.0648 \\
fandisk & 53.4734 & 1.0400 & 1.1499 & 0.0012 & 108.1353 & 1.2981 & 0.4451 & 0.0014 \\
flower & 142.6379 & 48.5886 & 54.8414 & 0.0580 & 286.3444 & 60.5215 & 24.6729 & 0.0664 \\
galera & 55.2452 & 0.7166 & 0.7792 & 0.0012 & 110.0765 & 0.8965 & 0.2512 & 0.0017 \\
happy & 49.9020 & 0.0024 & 0.0027 & 0.0000 & 100.2257 & 0.0030 & 0.0029 & 0.0000 \\
icosahedron & 49.7841 & 0.0437 & 0.0466 & 0.0002 & 98.9937 & 0.0545 & 0.0373 & 0.0003 \\
netsuke & 230.1374 & 159.0817 & 174.4530 & 0.2249 & 460.2821 & 198.9867 & 70.7092 & 0.3166 \\
pipe & 58.2976 & 1.1654 & 1.2764 & 0.0010 & 116.7092 & 1.4471 & 1.2878 & 0.0015 \\
pipe\_curve & 57.9949 & 1.1040 & 1.2031 & 0.0014 & 116.2264 & 1.3836 & 1.2681 & 0.0019 \\
star\_halfsmooth & 44.8887 & 0.1615 & 0.1800 & 0.0004 & 91.1262 & 0.2022 & 0.0891 & 0.0005 \\
star\_smooth & 46.4555 & 0.0831 & 0.0908 & 0.0003 & 93.8240 & 0.1047 & 0.0397 & 0.0004 \\
tortuga & 52.3346 & 0.1681 & 0.1937 & 0.0002 & 105.8191 & 0.2099 & 0.0967 & 0.0002 \\
yoda & 485.5565 & 367.5865 & 383.9735 & 1.4843 & 977.1352 & 458.9167 & 188.5749 & 9.4823 \\
\bottomrule
\end{tabular}}
\caption{ChamferL2 comparison under the spread noise settings \{40k, 100k, 200k, 500k\}.
\method{} achieves the lowest error on most datasets and remains stable even at the highest noise levels.
PointCVaR performs particularly well at the 100k noise level and can occasionally match or slightly outperform \method{} on a few datasets, but its performance drops noticeably as noise increases further.
Overall, \method{} is the most consistent and robust method across the full range of spread noise levels.}
\label{tab:spread_ChamferL2_2x2_levels}
\end{table*}

\subsection{Ablation Study}

\begin{figure}[h]
  \centering
    \includegraphics[width=0.6\linewidth]{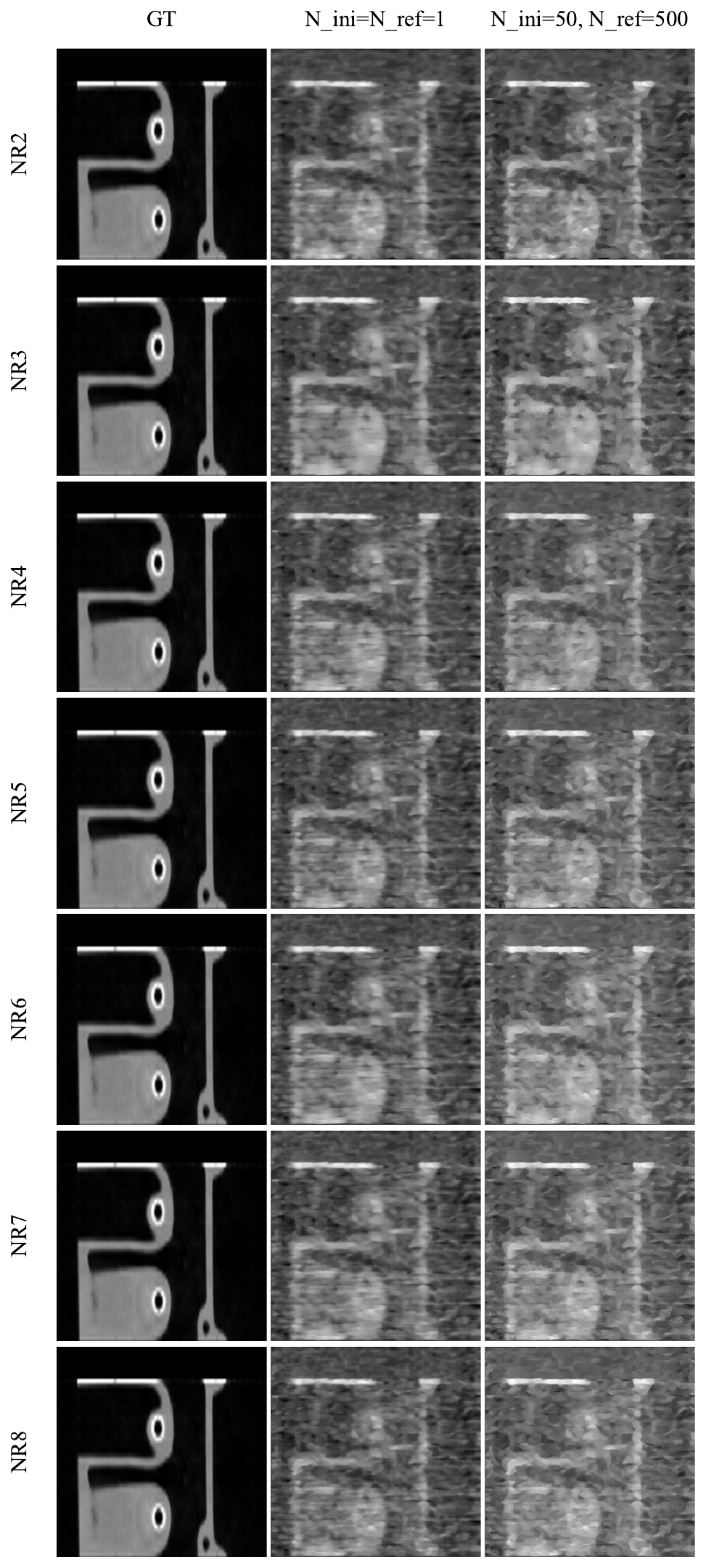}
    \caption{Sagittal slice views of reconstructed 3D volumes from the \textit{engine} dataset under low-SNR (\textit{P50}) conditions, illustrating the ablation study on the number of initialization steps, refinement iterations, and regions used in \method{}.}
  \vspace{-1.5em}
  \label{fig:ct_p50_slice_ablation_results_appendix}
\end{figure}

\Cref{fig:ct_p50_slice_ablation_results_appendix} and \Cref{tab:ablation_appendix} study how the number of regions and the number of initialization and refinement iterations influence reconstruction fidelity. The results show that increasing the number of initialization and refinement iterations does not improve reconstruction quality. In fact, the highest MS-SSIM and 3D PSNR values are consistently achieved with the smallest number of iterations.

Varying the number of regions also shows a clear trend: using 2–4 regions provides good  performance, while increasing the region count beyond this tends to introduce unnecessary model complexity without improving accuracy. Higher region counts slightly increase runtime (due to larger parameter sets).

Overall, the ablation confirms that a small number of iterations and a moderate number of regions (2–4) provides the best balance between reconstruction quality and computational efficiency.

\paragraph{Hyperparameter guidelines.}
Beyond the number of regions $M$, 3D FoJ is not highly sensitive to the remaining hyperparameters within reasonable ranges. 
The consistency weights $(\lambda_B,\lambda_C)$ are applied with a progressive schedule during refinement, making performance robust to their exact final values. 
The smoothness parameters $(\eta,\delta)$ operate in a stable regime, with $\eta \approx 10^{-2}$ providing sharp but numerically stable region transitions. 
Patch size and stride control the main bias--variance and runtime tradeoff: larger patches improve denoising at high noise levels but increase runtime and may over-regularize curved structures, while stride 2 provides a favorable efficiency/quality balance in our experiments.

\begin{table}[h]
\centering
\resizebox{\columnwidth}{!}{
\begin{tabular}{c c c c ccccc}
\toprule
\textbf{$\mathrm{N}_{\text{init}}$}&\textbf{$\mathrm{N}_{\text{refine}}$}&\textbf{\# regions}& \textbf{Batch size} & \textbf{Stride} & \textbf{Patch size} &  \textbf{MS-SSIM} & \textbf{3D PSNR} & \textbf{Runtime}\\
\midrule
1&1&2&6&2& 10 & 0.8048 & 17.025  & 15.2 min\\
1&1&3&6& 2& 10 & 0.7982 & 17.173   & 17.2 min\\
1&1&4&6&2 &  10 & 0.7975 & 16.701   & 20.3 min\\
1&1&5&6& 2&   10 & 0.7965 & 16.428   & 21.5 min\\
1&1&6&6& 2&   10 & 0.7897 & 16.502  & 22.8 min\\
1&1&7&6& 2&   10 & 0.7985 & 16.373   & 25.1 min \\
1&1&8&6& 2&   10 & 0.7969 & 16.124  & 27.7 min \\
\midrule
50&500&2&1& 2& 10 & 0.7423 &  16.361   & 952 min\\
50&500&3&1& 2& 10 & 0.7008 &  16.184  & 1188 min\\
50&500&4&1& 2&  10 & 0.7290 &  16.015   & 1322.4 min\\
50&500&5&1& 2&   10 & 0.7560 & 15.836   & 1456 min\\
50&500&6&1& 2&   10 & 0.7267 & 15.958 & 1588.8 min\\
50&500&7&1& 2&   10 & 0.7561 & 15.842 & 1707.5 min \\
50&500&8&1& 2&   10 & 0.7578 & 15.688 & 1870.5 min \\
\bottomrule
\end{tabular}}
\caption{Ablation study evaluating the effect of initialization steps, refinement iterations, and number of regions in \method{} on the \textit{engine} CT dataset under the highest noise level (\emph{P50}).}
\label{tab:ablation_appendix}
\end{table}

\subsection{Computational Cost and Curved-Structure Tradeoff}
\label{sec:app_compute_tradeoff}

We report additional runtime and peak-memory measurements for \method{} in \Cref{tab:runtime_memory_tradeoff}. 
The table includes both task-scale measurements for our default $M=3$ setting and the extra cost of using all $M=8$ sectors in a synthetic curved-tube experiment. 
All measurements are obtained on an NVIDIA RTX A6000 GPU with 48 GiB VRAM. 
The default $M=3$ setting is substantially more efficient, while using all $M=8$ sectors increases expressivity for high-curvature structures at higher memory and runtime cost.

\begin{table}[t]
  \centering
  \small
  \setlength{\tabcolsep}{4pt}
  \renewcommand{\arraystretch}{0.95}
  \begin{tabular}{lcc}
    \toprule
    Setting & Runtime & Peak memory \\
    \midrule
    Curved tube, $M=8/M=3$ avg. over all $R$ & $1.24\times$ & $1.82\times$ \\
    Curved tube, $M=8/M=3$ avg. for $R\geq6$ & $1.49\times$ & $1.82\times$ \\
    CT ($256^3$, 20 views), $M=3$ & 11.3 min & 44.5 GiB \\
    Cryo-ET ($1024\times1024\times300$), $M=3$ & 185.6 min & 33.2 GiB \\
    Point-cloud denoising ($256^3$), $M=3$ & 7.4 min & 33.2 GiB \\
    \bottomrule
  \end{tabular}
  \caption{\footnotesize Runtime and peak-memory measurements for \method{}. The first two rows report the relative cost of using all $M=8$ sectors instead of the default $M=3$ setting in the curved-tube experiment. The remaining rows report task-scale measurements for the default $M=3$ configuration.}
  \label{tab:runtime_memory_tradeoff}
\end{table}

\begin{figure}[t]
  \centering
  \includegraphics[width=\linewidth]{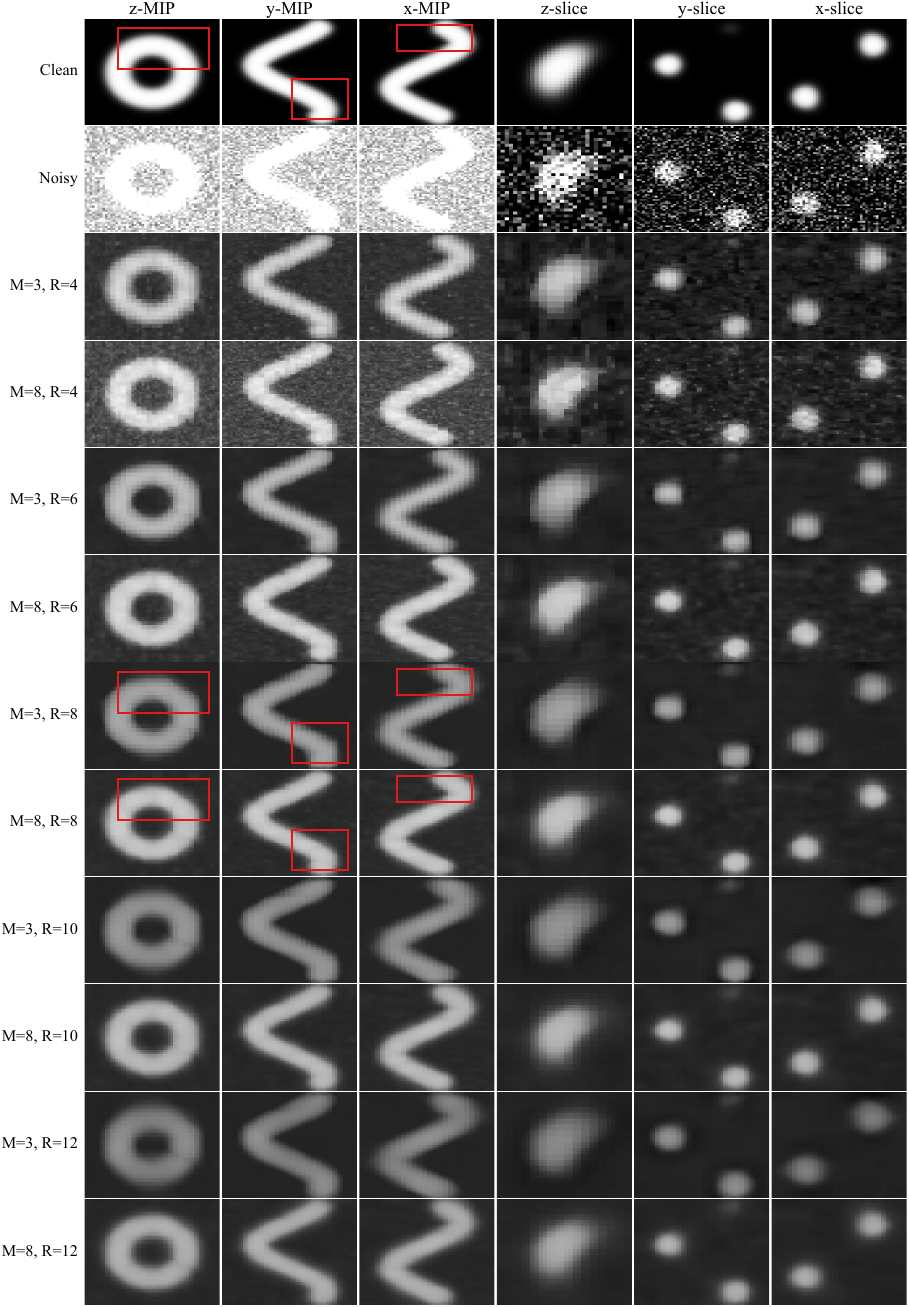}
  \caption{
  Curved-tube experiment on a synthetic $64^3$ volume. Columns show $z/y/x$
  maximum-intensity projections (MIPs) and central slices for the clean target,
  noisy input, and 3D FoJ outputs while sweeping local support size $R$ and active
  region count $M$. Red boxes mark representative curved regions for the clean
  target and the $R=8$ outputs. As $R$ grows, $M=3$ exhibits stronger local
  piecewise-planar smoothing, whereas $M=8$ better preserves curved structure at
  higher memory/runtime cost.
  }
  \label{fig:curved_tube_tradeoff}
\end{figure}

To connect this computational tradeoff to geometric expressivity, we evaluate a synthetic $64^3$ curved-tube denoising task. 
We sweep patch size $R\in\{4,6,8,10,12\}$ and compare the default $M=3$ setting against $M=8$, which uses all sectors induced by the three planes. 
As shown in \Cref{fig:curved_tube_tradeoff}, $M=3$ preserves curved geometry when the patch size is local relative to curvature. 
For larger patches or higher curvature, $M=8$ better preserves the curved structure, but at the increased memory and runtime cost reported in \Cref{tab:runtime_memory_tradeoff}.

\subsection{Convergence Behavior}
\label{sec:app_convergence}

\begin{figure}[t]
  \centering
  \includegraphics[width=0.85\linewidth]{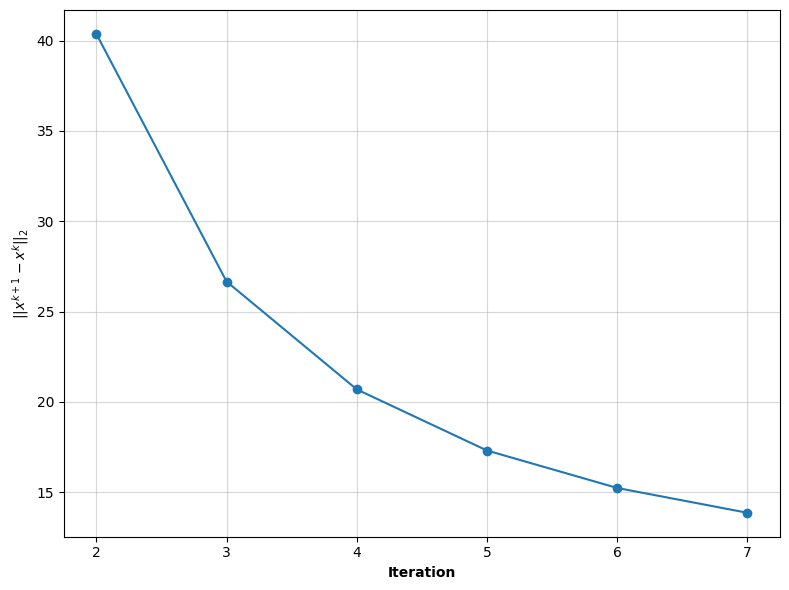}
  \caption{\footnotesize Empirical convergence behavior of the proximal reconstruction algorithm. The update magnitude $\|x^{k+1}-x^k\|_2$ decreases over iterations, showing stability despite the approximate single-step proximal update.}
  \label{fig:prox_convergence}
\end{figure}

We empirically analyze the convergence behavior of the proximal reconstruction procedure by plotting the update magnitude $\|x^{k+1}-x^k\|_2$ across outer iterations. 
As shown in \Cref{fig:prox_convergence}, the update norm decreases rapidly in early iterations and gradually stabilizes, which is consistent with convergence toward a fixed point despite using an inexact single-step FoJ proximal update.

\end{document}